\documentclass[lettersize,journal]{IEEEtran}
\usepackage[utf8]{inputenc}
\usepackage{mathrsfs}
\usepackage{amsmath,amsfonts,amssymb}
\usepackage{amsthm}
\usepackage{algorithmic}

\usepackage{array}
\usepackage{subfigure}
\usepackage[caption=false,font=normalsize,labelfont=sf,textfont=sf]{subfig}
\usepackage{textcomp}
\usepackage{stfloats}
\usepackage{url}
\usepackage{verbatim}
\usepackage{graphicx}
\usepackage{cite}
\usepackage{float}
\allowdisplaybreaks[4]
\begin{document}
\title{Convex Dual Theory Analysis of Two-Layer Convolutional Neural Networks with Soft-Thresholding }
\author{Chunyan Xiong, Mengli Lu, Xiaotong Yu, Jian Cao,  Zhong Chen, Di Guo, and Xiaobo Qu

\thanks{This work was
partially supported by National Natural Science Foundation
(62122064, 61971361 and 61871341), Natural Science Foundation of Fujian Province of China (2021J011184), President Fund of Xiamen University (20720220063), and the Xiamen
University Nanqiang Outstanding Talents Program.

Chunyan Xiong, Mengli Lu, Xiaotong Yu, Jian Cao, Zhong Chen, and Xiaobo Qu are with the School of Electronic Science and Engineering, Fujian Provincial Key Laboratory of Plasma and Magnetic Resonance, Biomedical Intelligent Cloud Research and Development Center, Xiamen University, Xiamen 361005, China.(Xiaobo Qu is the corresponding author with e-mail: quxiaobo@xmu.edu.cn)

Di Guo is with the School of Computer and Information Engineering, University of Technology, Xiamen 361024, China.}

\thanks{××××,××××,××××}}

\markboth{Journal of \LaTeX\ Class Files,~Vol.~14, No.~8, August~2021}%
{Shell \MakeLowercase{\textit{et al.}}: A Sample Article Using IEEEtran.cls for IEEE Journals}


\maketitle
\begin{abstract}
Soft-thresholding has been widely used in neural networks. Its basic network structure is a two-layer convolution neural network with soft-thresholding. Due to the network’s nature of nonlinearity and nonconvexity, the training process heavily depends on an appropriate initialization of network parameters, resulting in the difficulty of obtaining a globally optimal solution. To address this issue, a convex dual network is designed here. We theoretically analyze the network convexity and numerically confirm that the strong duality holds. This conclusion is further verified in the linear fitting and denoising experiments. This work provides a new way to convexify soft-thresholding neural networks.
\end{abstract}
 

\section{Introduction}

\IEEEPARstart{N}{eural} networks (NN) have been extensively employed in various  applications, including speech and image recognition \cite{machine1,machine2}, image classification\cite{machine3}, fast medical imaging\cite{yang2023physics} and biological spectrum reconstruction\cite{qu2020accelerated,huang2021exponential,27}, etc.
NN, however, is easy to stuck at the local optimum or the saddle point due to the network non-convexity (Fig. 1)\cite{ma2022diminishing}. This limitation prevents NN from reaching the global optimum \cite{liu2022convergence,liu2022construction,wang2020random}. To address this issue, proper initialization of network parameters is required in the training process \cite{kumar1,zhu2021gradinit,murgovski2013engine}. 
\begin{figure}
  \centering
  \includegraphics[width=2.6in]{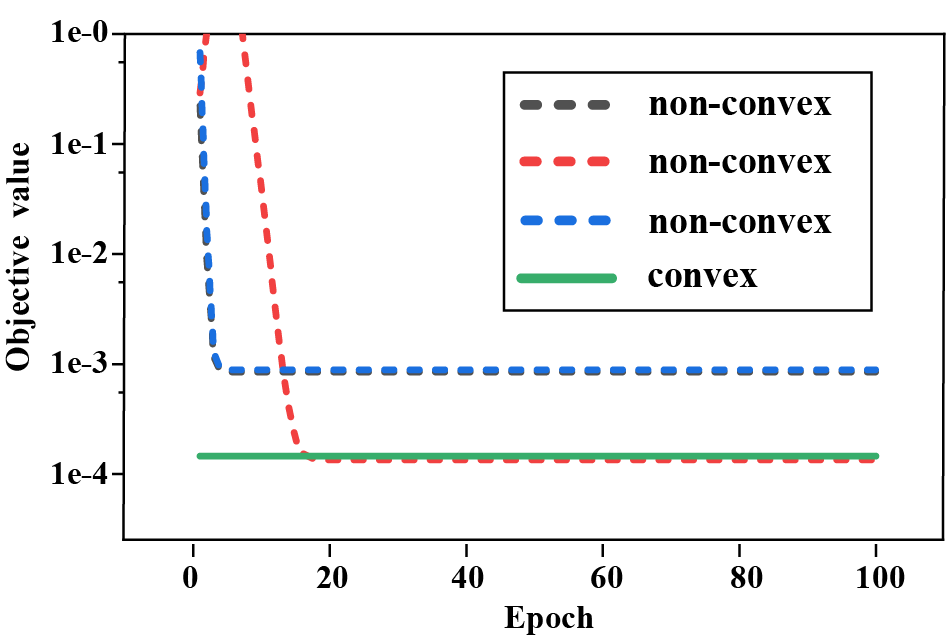}
 \caption{Toy example: In some bad cases, the non-convex neural network gets stuck in a local optimum or saddle point. The objective value of a two-layer non-convex and a convex neural network model for 1D vector fitting.
 The input of the network is $[-2,-1,0,1,2]^{T}$ and its ideal output (also called the label) is $[1,-1,-1,-1,1]^{T}$. Here, the bias term is included by concatenating a column of ones to the input.
 Under 3 random initialization trials of network parameters, the objective value of the non-convex neural network is different. Convex neural networks, which do not depend on the initialization, can be solved directly with a convex procedure to obtain the optimal value.  }
\end{figure}
\begin{figure*}
  \centering
  \includegraphics[width=6in]{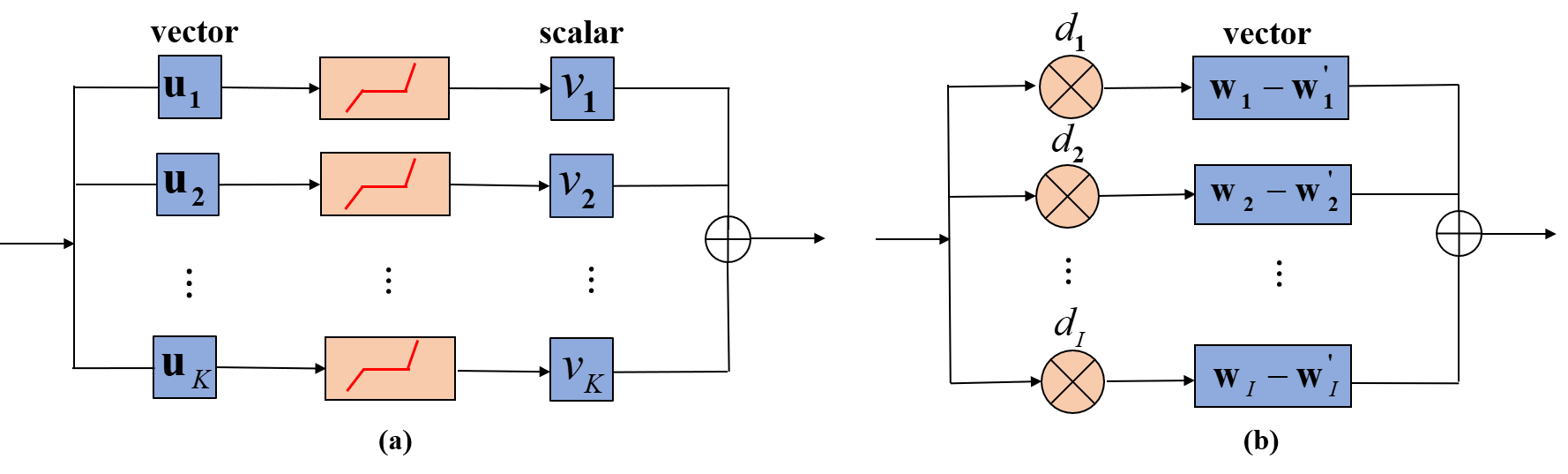}
  \caption{ Neural network structures. (a) Primal ST-CNN, (b) dual ST-CNN. }
\end{figure*}

Typical initialization strategies have been established\cite{machine3,glorot2010understanding,he2015delving} but the network may still encounter instability if the NN has multiple layers or branches\cite{zhu2021gradinit}. For example, the original Transformer model \cite{vaswani2017attention} did not converge without initializing the learning rate in a warm-up way \cite{xiong2020layer,huang2020improving,liu2020understanding}. Roberta \cite{liu2019roberta}  and GPT-3 \cite{brown2020language} had to tune the parameters of the optimizer ADAM\cite{Adam} for stability under the large batch size. Recent studies have shown that architecture-specific initialization can promote convergence  \cite{huang2020improving,zhang2019fixup,de2020batch,brock2021characterizing,brock2021high}. Even though, these initialization techniques hardly work to their advantage when conducting architecture searches, training networks with branching or heterogeneous components\cite{zhu2021gradinit}.

Convexifying neural networks is another way to make the solution not depend on initialization\cite{bengio2005convex},\cite{amos2017input}. 
At present, theoretical research on the convexification of NN focuses on finite-width networks which include fully connected networks\cite{Ergen2020convex,ergen2019convex,pilanci2020neural,wang2021convex,mishkin2022fast,sahiner2020vector,ergen2021global,ergen2021convex} and convolutional neural networks (CNN)\cite{22,ergen2020training}. The former is powerful to learn multi-level features\cite{fullconnected} but may require a large space and computational resources if the size of training data is large. The latter avoids this problem by reducing network complexity through local convolutions\cite{convolution0,convolution1,convolution2} and have been successfully applied in image processing\cite{2018Cruz,lawrence1997face,szegedy2015going,pinheiro2014recurrent}. Up to now, CNN has been utilized as an example in  convexifying networks under a common non-linear function, ReLU\cite{22,ergen2020training}.

To make the rest description clear, following previous theoretical work \cite{22}, we will adopt the denoising task handled by a basic two layers ReLU-CNN, for theoretical analysis of convexity.

Let $\widetilde{\textbf{X}}\in \mathbb{R}^{n\times h}$ denote a noise-free 2D image, $n$ and $h$ represent the width and height, respectively. $\widetilde{\textbf{X}}$ is contaminated by an additive noise $\textbf{E}$, whose 
entries are drawn from a probability distribution, such as $\mathscr{N}(0,\sigma^{2})$ in the case of i.i.d Gaussian noise. Then, the noisy observation $\widetilde{\textbf{Y}}$ is modeled as  $\widetilde{\textbf{Y}}=\widetilde{\textbf{X}}+\textbf{E}$.
Given a set of convolution filters, $\widetilde{\textbf{U}}_{k}\in\mathbb{R}^{m\times m}(k=1,...,K)$, noise-suppressed images are obtained under each filter and then linearly combined according to\cite{22}
\begin{align}
\sum_{k=1}^{K} 
\left( \widetilde{\textbf{Y}} \otimes\widetilde{\textbf{U}}_{k}\right)_{+}\otimes \widetilde{v}_{k}, \end{align}
where $\otimes$ represents the 2D convolution operation, $(\cdot)_{+}$ denotes an element-wise ReLU operation, and $\widetilde v_{k}\in\mathbb{R}$ is a 1×1 kernel used as the weight in the linear combination. Then, convolution kernels, i.e. $\widetilde{\textbf{U}}_{k}$ and $v_{k}$, are obtained by minimizing the prediction loss between the noise-free image and the network output as
\begin{align}
\min_{\widetilde{\textbf{U}}_{k},\widetilde{v}_{k}}
\left \| \sum_{k=1}^{K} 
\left( \widetilde{\textbf{Y}} \otimes\widetilde{\textbf{U}}_{k}\right)_{+}\otimes \widetilde{v}_{k}-\widetilde{\textbf{X}}\right\|_{F}^{2}.\end{align}

To  reduce the network complexity, Eq. (2) is further improved to 
an object value as\cite{22}
\begin{align}
&\min_{\widetilde{\textbf{U}}_{k},\widetilde{v}_{k}}
\left \| \sum_{k=1}^{K} 
\left( \widetilde{\textbf{Y}} \otimes\widetilde{\textbf{U}}_{k}\right)_{+}\otimes \widetilde{v}_{k}-\widetilde{\textbf{X}}\right\|_{F}^{2}\\&
+\beta\sum_{k=1}^{K}(\left \| \widetilde{\textbf{U}}_{k}\right \|_{F}^{2}+|\widetilde{v}_{k}|^{2}),\nonumber \end{align}
by constraining the energy (or the power of norm) of all convolution kernels. The $\beta>0$ is a hyper-parameter to trade the prediction loss with the convolution kernel energy. 
   
To convexify the primal network in Eq. (3), the convex duality theory was introduced to convert Eq. (3) into a dual form, enabling the reach of global minimum\cite{22}. No gap between the primal and dual objective values has been demonstrated theoretically and experimentally\cite{22}. This work inspired us to convexify other networks, for example, replacing ReLU with soft-thresholding.

Soft-thresholding (ST) is another non-linear function that has been widely adopted in CNN\cite{23,24,25,26,27}. To the best of our knowledge, the convex form of primal Two-Layer Convolutional Neural Networks with Soft-Thresholding (primal ST-CNN) has not been set up. This work is to design its convex structure and provide the theoretical analysis (Fig. 2). First, we derive the weak duality of the primal ST-CNN using the Lagrange dual theory. Second, the nonlinear operation is converted into a linear operation (Fig. 3), which is used to divide hyperplanes and provide exact representations in the training process. Third, we theoretically prove that the strong duality holds between the two-layer primal ST-CNN (Fig. 2(a)) and its dual form (dual ST-CNN) (Fig. 2(b)). Finally, experiments are conducted to support theoretical findings.

 \begin{figure}
  \centering
  \includegraphics[width=2.5in]{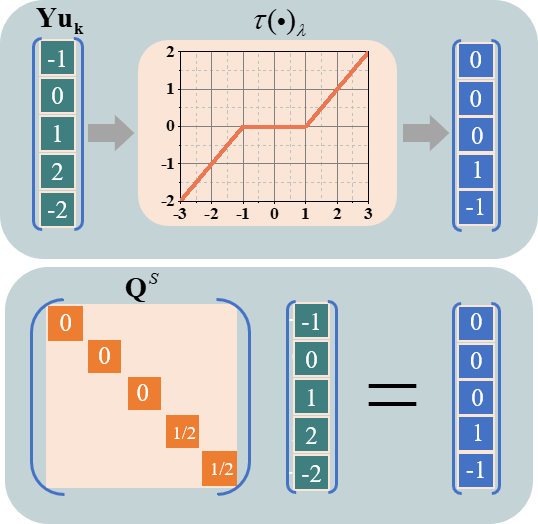}
  \caption{Toy example: Converting non-linear operation to linear operation. (a) Soft-thresholding in the primal ST-CNN, (b) diagonal matrix in the dual ST-CNN. }
\end{figure}

The rest of this paper is organized as follows: Section \uppercase\expandafter{\romannumeral2} introduces preliminaries. Section \uppercase\expandafter{\romannumeral3} presents the main theorem. Section \uppercase\expandafter{\romannumeral4} shows experimental results and Section  \uppercase\expandafter{\romannumeral5} makes the conclusion.    

\section{Preliminaries}
\subsection{Notations}

Matrices and vectors are denoted by uppercase and lowercase bold letters, respectively. $\left \| \cdot \right \|_{2} $ and $\left \| \cdot \right \|_{F}$ represents Euclidean and Frobenius norms, respectively.
We partition $P_{s}\subset\mathbb{R}^{m^{2}}$ into the following subsets
\begin{align}P_{s}=l_{1}\cup l_{2}\cup l_{3},\end{align}
where 
\begin{align} &l_{1}=\{\textbf{u}_{k}|~\textbf{y}_{i}^{T}\textbf{u}_{k}\le -\lambda\}, ~~~~~~~H_{1}=\{i|~\textbf{y}_{i}^{T}\textbf{u}_{k}\le -\lambda\},\\\nonumber
&l_{2}=\{\textbf{u}_{k}|~-\lambda \le \textbf{y}_{i}^{T}\textbf{u}_{k}\le \lambda\},H_{2}=\{i|~-\lambda \le \textbf{y}_{i}^{T}\textbf{u}_{k}\le \lambda\},\\\nonumber
&l_{3}=\{\textbf{u}_{k}|~\textbf{y}_{i}^{T}\textbf{u}_{k}\ge \lambda\},~~~~~~~~~ H_{3}=\{i|~\textbf{y}_{i}^{T}\textbf{u}_{k}\ge \lambda\},\\\nonumber
&\{\textbf{y}_{i}\in{\mathbb{R}^{m^{2}}\}}_{i=1}^{I}, ~I=nh, ~~
\textbf{u}_{k}\in{\mathbb{R}^{m^{2}}},~~\lambda\in\mathbb{R} .  
\end{align}

We denote \begin{align}S=S_{1}\cup S_{2}\cup S_{3},\end{align}
where
\begin{align}
&S_{1}=\{i|~i\in H_{1}\}\cup \{i|~i\in H_{2}\},\\\nonumber
&S_{2}=\{i|~i\in H_{2}\},\nonumber\\
&S_{3}=\{i|~i\in H_{2}\}\cup \{i|~i\in H_{3}\}.\nonumber
\end{align}

$$\textbf{Y}=[\textbf{y}_{1},\textbf{y}_{2},...,\textbf{y}_{I}]^{\top}\in{\mathbb{R}^{I\times m^{2}}}, ~~\textbf{Q}^{S}\in {\mathbb{R}^{I\times I}} ~\text{is a}$$
diagonal matrix, its diagonal elements are as follows

\begin{align}\textbf{Q}_{ii}=\left\{\begin{matrix}
  &\\\frac{\textbf{y}_{i}^{T}\textbf{u}_{k} +\lambda }{\textbf{y}_{i}^{T}\textbf{u}_{k}}, \quad \text{if} \quad i\in S_{1},\\
  &\\\quad 0,\quad  \quad \text{if} \quad i\in S_{2},\\
  &\\\frac{\textbf{y}_{i}^{T}\textbf{u}_{k} -\lambda }{\textbf{y}_{i}^{T}\textbf{u}_{k}},\quad \text{if} \quad i\in S_{3}.
\end{matrix}\right.
\end{align}
\begin{align}\textbf{Q}^{S}=\textbf{Q}^{S_{1}}+\textbf{Q}^{S_{2}}+\textbf{Q}^{S_{3}},\end{align}

We denote \begin{align}\nonumber
&\textbf{Q}^{S} ~\text{in}~ \text{Q}^{S}\textbf{Y}\textbf{u}_{k}\ge0 ~\text{as}~ \textbf{Q}^{S_{1}},\\\nonumber
&\textbf{Q}^{S} ~\text{in}~ \textbf{Q}^{S}\textbf{Y}\textbf{u}_{k}=0 ~\text{as}~ \textbf{Q}^{S_{2}},\\\nonumber
&\textbf{Q}^{S} ~\text{in}~ \textbf{Q}^{S}\textbf{Y}\textbf{u}_{k}\le 0 ~\text{as}~ \textbf{Q}^{S_{2}}.\nonumber
\end{align}

\begin{align}P_{S}=\{\textbf{u}_{k}|~P_{S_{1}}\cup P_{S_{2}}\cup
P_{S_{3}}\},\end{align}
where \begin{align}&P_{S_{1}}=\{\textbf{u}_{k}|~\textbf{Q}^{S_{1}}\textbf{Y}\textbf{u}_{k}\ge0 ,\quad \forall i\in S_{1}\},\\
&P_{S_{2}}=\{\textbf{u}_{k}|~\textbf{Q}^{S_{2}}\textbf{Y}\textbf{u}_{k}=0,\quad \forall i\in S_{2}\},\nonumber\\
&P_{S_{3}}=\{\textbf{u}_{k}|~\textbf{Q}^{S_{3}}\textbf{Y}\textbf{u}_{k}\le 0, \quad \forall i\in S_{3}\}.\nonumber
\end{align}

\subsection{Basic Lemmas and Definitions}
\emph{Lemma 1} (Slater's condition\cite{Boyd}):
Consider the optimization problem
\begin{align}
&~~~~~~~~~~~~~\min_{\textbf{x}} f_{0}(\textbf{x})\\\nonumber
&~~~~~~~~~~~~~~\text{s.t.}~f_{j}(\textbf{x})<0,\quad j=1,...,J,\quad \textbf{Ax}=\textbf{b},
\end{align}
where~$\textbf{A}\in \mathbb{R}^{m\times n}$,~$ \textbf{x}\in \mathbb{R}^{n}$,~$\textbf{b}\in \mathbb{R}^{m}$,
$f_{0},...,f_{J}$ are convex functions.

If there exists an $\textbf{x}^{*}\in \textbf{relint} D$ 
(where $\textbf{relint}$ denotes the relative interior of the convex set $D:= \cap_{j=0}^{J}\text{dom} (f_{j})$), such that 
\begin{align} f_{j}(\textbf{x}^{*})<0,\quad j=0,...,J,\quad \textbf{A}\textbf{x}^{*}=\textbf{b}.\end{align}

Such a point is called strictly feasible since the inequality constraints hold with strict inequalities. The strong duality holds if Slater's condition holds (and the problem is convex).

\emph{Lemma 2} (Sion's Minimax theorem \cite{sion1958general},\cite{kindler2005simple}): Let $X$ and $Y$ be nonvoid convex and compact subsets of two linear topological spaces, and let $f: X\times Y\to \mathbb{R}$ be a function that is upper semicontinuous and quasi-concave in the first variable and lower semicontinuous and quasi-convex in the second variable. Then 
\begin{align}
\min_{y\in{Y}}\max_{x\in{X}}f(x,y) = \max_{x\in{X}}\min_{y\in{Y}}f(x,y).
\end {align}

\emph{Lemma 3} (Semi-infinite programming \cite{36}): Semi-infinite programming problems of the form 
\begin{align}
\min_{\textbf{x}\in\mathbb{R}^{n}}f(\textbf{x}) ~\text{subject to} ~g(\textbf{x},w) \le 0, w\in\Omega,
\end{align}
where $\Omega$ is a (possibly infinite) index set, $\mathbb{\overline{R}}=\mathbb{R}\cup\{{+\infty}\}\cup\{{-\infty}\}$ denotes the extended real line, $f:\mathbb{R}^{n}\to\mathbb{\overline{R}} $ and $g:\mathbb{R}^{n}\times \Omega\to\mathbb{R}$. The above optimization problem is performed in the finite-dimensional space $\mathbb{R}$ and, if the index set $\Omega$ is infinite, is subject to an infinite number of constraints, therefore, it is referred to as a semi-infinite programming problem. 

\emph{Lemma 4} (An extension of Zaslavsky's hyperplane arrangement theory\cite{37}): Consider a deep rectifier network with $L$ layers, $n_{l}$ rectified linear  units at each layer $l$, and an input of dimension $n_{0}$. The maximal number of regions of this neural network is at most
\begin{align}\sum_{(j_{1},...,j_{L})}\prod_{l=1}^{L}\left ( \begin{matrix}n_{l}
 \\j_{l}
\end{matrix} \right ),\end{align}
where $J=\{(j_{1},...,j_{L})\in \mathbb{Z}^{L}: 0\le j_l\le \text{min}\{n_{0},n_{1}\!-\!j_{1},...,$\\$n_{l-1}\!-\!j_{l-1},
n_{l}\}~\forall~l=1,...,L\}.$ \text{This bound is tight when} $L=1$.

\emph{Definition 1} (Optimal duality gap\cite{Boyd}): The optimal value of the Lagrange dual problem is denoted as $d^{\ast }$, and the optimal value of the primal problem is denoted as $p^{\ast }$. The weak duality is defined as $d^{\ast }$ is the best lower bound of $p^{*}$ as follows
\begin{align}d^{\ast }\le p^{\ast}.\end{align}

 The difference $p^{\ast}-d^{\ast }$ is called the optimal duality gap of the primal problem.
 
\emph{Definition 2} (Zero duality gap \cite{Boyd}): If the equality 
\begin{align} d^{\ast}=p^{\ast}\end{align} holds, i.e. the optimal duality gap is zero, then we say that the strong duality holds. 
Strong duality means that a best bound, which can be obtained from the Lagrange dual function, is tight.

\emph{Definition 3} (Hyperplanes and halfspaces\cite{Boyd}):

A hyperplane is a set of the form 
$\{\textbf{x}|~\textbf{a}^{T}\textbf{x}=b\}$
where $\textbf{a}\in \mathbb{R}^{n},~ \textbf{x}\in \mathbb{R}^{n\times 1},~ \textbf{a}\ne 0$ and $b\in \mathbb{R}$.

A hyperplane divides $\mathbb{R}^{n}$ into halfspaces. A halfspace is a set of the form 
$\{\textbf{x}|~\textbf{a}^{T}\textbf{x}\le b\}$ where $\textbf{a}\ne 0$, i.e. the solution set of one (nontrivial) linear inequality. This is illustrated in Fig. 4.

\begin{figure}
  \centering
  \includegraphics[width=2.5in]{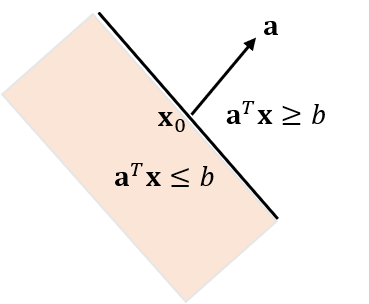}
  \caption{A hyperplane defined by $\textbf{a}^{\top}\textbf{x}=b$ in $\mathbb{R}^{2}$ determines two halfspaces. The halfspace determined by $\textbf{a}^{\top}\textbf{x}\ge b$ is the halfspace extending in the direction $\textbf{a}$. The halfspace determined by $\textbf{a}^{\top}\textbf{x}\le  b$ extends in the direction $-\textbf{a}$. The vector $\textbf{a}$ is the outward of this halfspace. }
  
\end{figure}

\begin{figure}
  \centering
  \includegraphics[width=3.4in]{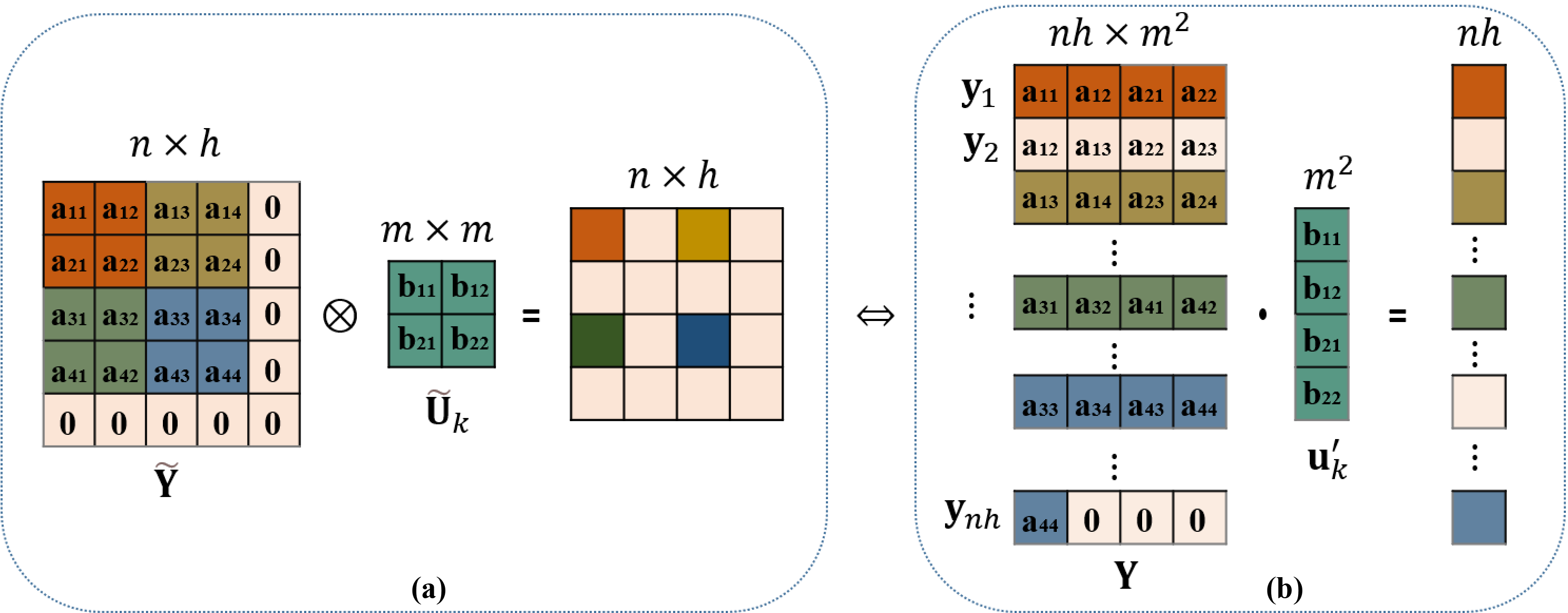}
  \caption{Replacing convolutional operations with matrix multiplication. (a) Convolution in Eq. (19), (b) matrix multiplication in Eq. (20).}
\end{figure}

\section{Model and Theory}
\subsection{Proposed Model}
A two-layer primal ST-CNN is expressed as follows
\begin{align}
p^{\ast}=&\min_{\widetilde{\textbf{U}}_{k},\widetilde{v}_{k}}
\left \| \sum_{k=1}^{K} \tau \left( \widetilde{\textbf{Y}} \otimes\widetilde{\textbf{U}}_{k}\right) _{\lambda}\otimes \widetilde{v}_{k}- \widetilde{\textbf{X}}\right\|_{F}^{2}\nonumber
\\&+\beta\sum_{k=1}^{K}(\left \| \widetilde{\textbf{U}}_{k}\right \|_{F}^{2}+|\widetilde{v}_{k}|^{2}), \end{align}
where the main difference between Eq. (19) and Eq. (3) is an element-wise soft-thresholding operator $\tau(a_{ij})_{\lambda}=(|a_{ij}|-\lambda)_{+}sign(a_{ij})$,
 $\otimes$ represents the 2D convolution operation, $\widetilde{\textbf{Y}}\in \mathbb{R}^{n\times h}$ is the input,
$\widetilde{\textbf{U}}_{k}\in \mathbb{R}^{m\times m}$, $\widetilde v_{k}\in\mathbb{R}$, $\beta>0$.

Replacing convolutional operations with matrix multiplication (Fig. 5), Eq. (19) can be converted into the following form
\begin{align}
p^{\ast}=&\min_{\textbf{u}'_{k},v'_{k}}
\left\|\sum_{k=1}^{K} \tau  \left (\textbf{Y}\textbf{u}'_{k}\right)_{\lambda }  v'_{k}-\textbf{x}\right\|_{2}^{2}\nonumber\\
&+\beta \sum_{k=1}^{K}(\left \| \textbf{u}'_{k} \right \|_{2}^{2}+|v'_{k}|^{2}),
\end{align}
where $\textbf{Y}=[\textbf{y}_{1},\textbf{y}_{2},...,\textbf{y}_{I}]^{\top}\in{\mathbb{R}^{I\times m^{2}}}$ is the input, $I=nh,$ $\{\textbf{y}_{i}\in{\mathbb{R}^{m^{2}}\}}_{i=1}^{I}$,  $\textbf{x}\in \mathbb{R}^{I}$ is the label, $\textbf{u}'_{k}\in{\mathbb{R}^{m^{2}}}, v'_{k}\in\mathbb{R}$.

Next, we introduce the main theory (Theorem 1) that converts a two-layer primal ST-CNN (Fig. 6(a)) into a convex dual ST-CNN (Fig. 6(b)).

\begin{figure*}
  \centering
  \includegraphics[width=6in]{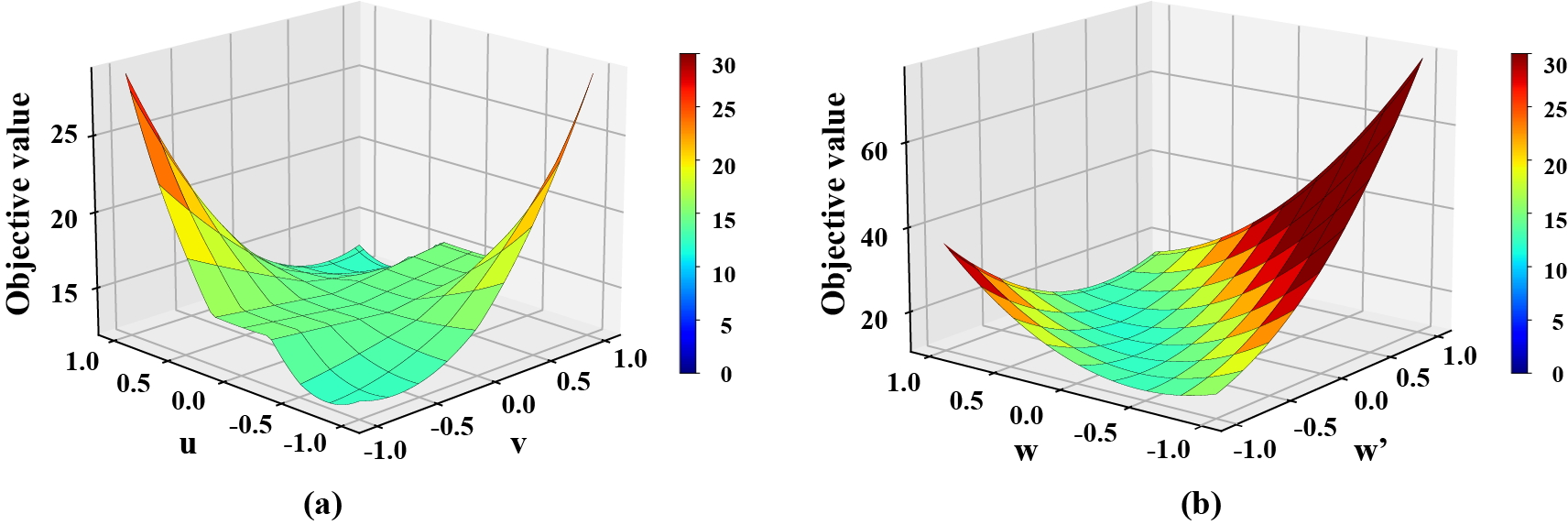}
  \caption{Objective value of a two-layer primal ST-CNN and dual ST-CNN trained with ADAM on a one-dimensional dataset. 
  Assuming $\textbf{x}=[-1,2,0,1,2]$ and $\textbf{y}=[2,1,2,1,2]$, which are the input and output, respectively. (a) Non-convex primal ST-CNN, (b) convex dual ST-CNN.}
\end{figure*}
\begin{figure}
  \centering
  \includegraphics[width=2.5in]{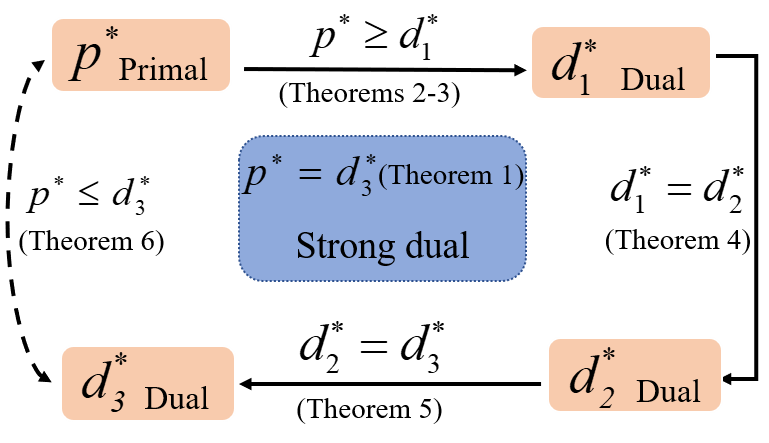}
  \caption{The main derivation process.}
\end{figure}

\subsection{Theoretical Analysis}
\emph{Theorem 1} (Main theory): There exists $k^{*}\le I$ such that if the number of convolution filters $k\ge (k^{*}+1)$, a two-layer ST-CNN (Eq. 20) has a strong duality satisfy form. This form is given through finite-dimensional convex programming as
\begin{align}
d_{3}^{*}=\min_{\textbf{w} _{i}\in 
p_{w},\textbf{w}'_{i}\in p_{w'}}&\left \| \sum_{i=1}^{I}\textbf{Q}^{S}\textbf{Y}(\textbf{w}'_{i}-\textbf{w}_{i})-\textbf{x}\right \|_{2}^{2}\\\nonumber
&+2\beta\sum_{i=1}^{I} (\left \| \textbf{w}'_{i} \right \|_{2}+ \left \|\textbf{w}_{i} \right \|_{2}),\end{align} 
where  $\textbf{Q}^{S}$ is a diagonal matrix, and its diagonal elements for $\textbf{Q}_{ii}$ take the following values
\begin{align}\textbf{Q}_{ii}=\left\{\begin{matrix}
  & \\\frac{\textbf{y}_{i}^{T}\textbf{u}_{k} +\lambda }{\textbf{y}_{i}^{T}\textbf{u}_{k}}, \quad \text{if} \quad i\in S_{1},\\
  & \\\quad 0,\quad  \quad \text{if} \quad i\in S_{2},\\
  & \\\frac{\textbf{y}_{i}^{T}\textbf{u}_{k} -\lambda }{\textbf{y}_{i}^{T}\textbf{u}_{k}},\quad \text{if} \quad i\in S_{3}.
\end{matrix}\right.\end{align}
$\textbf{Y}=[\textbf{y}_{1},\textbf{y}_{2},...,\textbf{y}_{I}]^{\top}\in{\mathbb{R}^{I\times m^{2}}}$, 
$\textbf{w}_{i}$ and $\textbf{w}_{i}'$ are both dual variables, and they correspond to $\textbf{u}'_{k}$ and $v_{k}'$ in Eq. (20) which are learnable parameters.
$\textbf{x}\in\mathbb{R}^{I}$ is the label.
\begin{align}&p_{w}=\left \{ \textbf{w}_{i}|\textbf{Q}^{S_{1}}\textbf{Y}\textbf{w}_{i}\ge 0,\textbf{Q}^{S_{2}}\textbf{Y}\textbf{w}_{i} =0,\textbf{Q}^{S_{3}}\textbf{Y}\textbf{w}_{i}\le  0\right \},\\
&p_{w'}=\left \{ \textbf{w}'_{i}|\textbf{Q}^{S_{1}}\textbf{Y}\textbf{w}'_{i}\ge 0,\textbf{Q}^{S_{2}}\textbf{Y}\textbf{w}'_{i}=0,\textbf{Q}^{S_{3}}\textbf{Y}\textbf{w}'_{i}\le  0\right \}.\nonumber\end{align}

\begin{align}\textbf{Q}^{S}=\textbf{Q}^{S_{1}}+\textbf{Q}^{S_{2}}+\textbf{Q}^{S_{3}}.\end{align}

\emph{Remark:}
The constraints on  $\textbf{w}$ and $\textbf{w}'$ in $p_{w}$ and $p_{w}'$ arise from the segmentation property of the soft thresholding. We first randomly generate the vector $\bar{\textbf{w}}$ to do convolution with the input $\textbf{Y}$ and generate the corresponding $\textbf{Q}^{S}$  based on the value of $\textbf{Y}\bar{\textbf{w}}$. Then we input $\textbf{Y}$ and $\textbf{Q}^{S}$ into our dual ST-CNN, and Eq. (21) is used in our objective function (objective loss). Because it is an objective function with constraints $p_{w}$ and $p_{w}'$, hence, we use hinge loss (adding constraints to the objective function) as the loss function in experiments. There exist $\textbf{w}_{i}$, $\textbf{w}'_{i}$ such that $\bar{\textbf{w}}=\textbf{w}'_{i}-\textbf{w}_{i}$.

Before proving the main theory (Theorem 1), we present the following main derivation framework (Fig. 7).

1) Theorem 2: Scaling $\left \| \textbf{u}'_{k} \right \|_{2}^{2}+|v'_{k}|^{2}$ in the primal ST-CNN (Eq. 20);

2) Theorem 3: Eliminating variables to obtain an equivalent convex optimization model under the principle of Lagrangian dual theory;

3) Theorem 4: Convert nonlinear operations to linear operations using a diagonal matrix;

4) Theorem 5: Exact representation of a two-layer ST-CNN;

5) Theorem 6: Prove zero dual gaps (strong duality).

\emph{Theorem 2}: To scaling $\textbf{u}'_{k}, v'_{k} $, let $\textbf{u}_{k}=\varepsilon \textbf{u}'_{k}$,\quad
$v_{k}=\frac{1}{\varepsilon }{v'}_{k}$, \begin{align}
p^{\ast}=&\min_{\textbf{u}'_{k},v'_{k}}
\left\|\sum_{k=1}^{K} \tau  \left (\textbf{Y}\textbf{u}'_{k}\right)_{\lambda }  v'_{k}-\textbf{x}\right\|_{2}^{2}\nonumber\\
&+\beta \sum_{k=1}^{K}(\left \| \textbf{u}'_{k} \right \|_{2}^{2}+|v'_{k}|^{2}),
\end{align} the primal ST-CNN can be translated as
\begin{align}
p^{\ast}=&\min_{{\left \| \textbf{u} _{k} \right \|_{2}\le 1}}\min_{v_{k}\in\mathbb{R}}
\left \| \sum_{k=1}^{K} \tau \left (\textbf{Y}\textbf{u} _{k}\right )_{\lambda} v_{k}-\textbf{x}  \right\|_{2}^{2}\nonumber\\&+2\beta \sum_{k=1}^{K}(|v_{k}|),
\end{align}

where $\varepsilon$ is introduced so that the scaling has no effect on the network output, the proof of Theorem 2 is provided in Appendix A. 

Then, according to Eq. (26), we can obtain an equivalent convex optimization model by using the Lagrangian dual theory.

\emph{Theorem 3}: \begin{align}\nonumber
p^{\ast}=&\min_{{\left \| \textbf{u} _{k} \right \|_{2}\le 1}}\min_{v_{k}\in\mathbb{R}}
\left \| \sum_{k=1}^{K} \tau\left ( \textbf{Y}\textbf{u} _{k}\right )_{\lambda }v_{k}-\textbf{x} \right\|_{2}^{2}\\\nonumber&+2\beta \sum_{k=1}^{K}|{v} _{k}|.\nonumber\end{align}
 is equivalent to 
\begin{align}d_{1}^{\ast}&=\max_{\left \| \textbf{u}_{k} \right \|_{2}\le 1, \textbf{z}:|\textbf{z}^{T}\tau  \left ( \textbf{Y}\textbf{u}_{k}\right )_{\lambda } |\le 2\beta}-\frac{1}{4}\left \|\textbf{z}-2\textbf{x} \right \|_{2}^{2}+ \left \|\textbf{x}\right \| _{2}^{2}.\end{align}

Proof: By reparameterizing the problem, let 
\begin{align}
\textbf{r}=\sum_{k=1}^{K} \tau \left (\textbf{Y}\textbf{u} _{k}\right )_{\lambda} v_{k}-\textbf{x},
\end{align}

where $\textbf{r}\in {\mathbb{R}^{I}}$, hence, we have
\begin{align}
&d_{1}^{\ast}=\min_{\left \| \textbf{u}_{k} \right \|_{2}\le 1 }\min_{{v}_{k},\textbf{r}}\left \|\textbf{r}\right\|_{2}^{2}+2\beta\sum_{k=1}^{K}|{v} _{k}|,\nonumber\\&s.t.\quad \textbf{r}=\sum_{k=1}^{K} \tau \left (\textbf{Y}\textbf{u} _{k}\right )_{\lambda} v_{k}-\textbf{x}.
\end{align}

Introducing the Lagrangian variable $\textbf{z}$, and $\textbf{z}\in \mathbb{R}^{I}$,~$\textbf{z}^{T} \in \mathbb{R}^{1\times I}$, and obtaining the Lagrangian dual form of the primal ST-CNN as follows
\begin{align}
d_{1}^{\ast}=&\min_{\left \| \textbf{u} _{k} \right \|_{2}\le 1}\min_{v_{k},\textbf{r}}\max_{\textbf{z}}
\left \| \textbf{r}  \right\|_{2}^{2}+2\beta\sum_{k=1}^{K} |v _{k}|+\textbf{z}^{T}\textbf{r}\nonumber\\&+\textbf{z}^{T}\textbf{x}-\textbf{z}^{T}\sum_{k=1}^{K} \tau \left(\textbf{Y}\textbf{u}_{k}\right)_{\lambda} v_{k}.
\end{align}

Using Sion's minimax theorem\cite{sion1958general},\cite{kindler2005simple} to change the order of maximum and minimum
\begin{align}
d_{1}^{\ast}=&\min_{\left \| \textbf{u} _{k} \right \|_{2}\le 1}\max_{\textbf{z}}\min_{v_{k},\textbf{r}}
\left \| \textbf{r}  \right\|_{2}^{2}+2\beta\sum_{k=1}^{K} |{v} _{k}|+\textbf{z}^{T}\textbf{r}\nonumber\\&+\textbf{z}^{T}\textbf{x}-\textbf{z}^{T}\sum_{k=1}^{K} \tau \left ( \textbf{Y}\textbf{u} _{k}\right ) _{\lambda }v_{k}.
\end{align}

Minimizing the objective function Eq. (31) with $\textbf{r}$ as a variable
\begin{align}
\left \| \textbf{r} \right \|_{2}^{2}+\textbf{z}^{T}\textbf{r}=\left\|\textbf{r}+\frac{1}{2}\textbf{z}\right\|_{2}^{2}-\frac{1}{4}\left\|\textbf{z}\right\|_{2}^{2}.
\end{align}

When $\textbf{r}=-\frac{1}{2}\textbf{z}$, Eq. (31) takes the optimal value. Hence, Eq. (31) can be translated to
\begin{align}
d_{1}^{\ast}=&\min_{\left\|{\textbf{u} _{k}}\right\|_{2}\le 1}\max_{\textbf{z}}\min_{v_k{}}
 -\frac{1}{4}\left\|\textbf{z} \right\|_{2}^{2}+2\beta \sum_{k=1}^{K}| {v} _{k}|+\textbf{z}^{T}\textbf{x}\nonumber\\&-\textbf{z}^{T}\sum_{k=1}^{K} \tau  \left ( \textbf{Y}\textbf{u} _{k}\right )_{\lambda } v_{k}.\end{align}

Let
 \begin{align}f=\min_{v_{k}}2\beta \sum_{k=1}^{K}\left |v_ {k}\right |  -\textbf{z}^{T}\sum_{k=1}^{K}\tau  \left ( \textbf{Y}\textbf{u} _{k}\right)_{\lambda } v_{k}, \end{align}
 eliminating the variable $v_{k}$ in the primal ST-CNN, hence
\begin{align}
\max_{\textbf{z}:\left \| \textbf{u}_{k} \right \|_{2}\le 1}|\textbf{z}^{T}\sum_{k=1}^{K}\tau \left ( \textbf{Y}\textbf{u} _{k}\right ) _{\lambda } |\le 2\beta.
\end{align}

Eq. (33) is equivalent to the following optimization problem
\begin{align}
d_{1}^{\ast}&=\max_{\left \| \textbf{u}_{k} \right \|_{2}\le 1, \textbf{z}:|\textbf{z}^{T}\sum_{k=1}^{K}\tau  \left ( \textbf{Y}\textbf{u} _{k}\right )_{\lambda } |\le 2\beta}-\frac{1}{4}\left \|\textbf{z}-2\textbf{x} \right \|_{2}^{2}+ \left \|\textbf{x}\right \| _{2}^{2}.
\end{align}
$\hfill\qedsymbol$

Next, to divide hyperplanes and provide an exact representation, we convert the nonlinear operation $\tau(\cdot)_{\lambda }$ into  the linear operator using the diagonal matrix $\textbf{Q}^{S}$.

\emph{Theorem 4}:  \begin{align}
d_{1}^{\ast}&=\max_{\left \| \textbf{u}_{k} \right \|_{2}\le 1, \textbf{z}:|\textbf{z}^{T}\sum_{k=1}^{K}\tau  \left ( \textbf{Y}\textbf{u} _{k}\right )_{\lambda } |\le 2\beta}-\frac{1}{4}\left \|\textbf{z}-2\textbf{x} \right \|_{2}^{2}+ \left \|\textbf{x}\right \| _{2}^{2},
\nonumber\end{align} can be represented as a standard finite-dimensional program
\begin{align}
d_{2}^{\ast}=\max_{\textbf{z}}-\frac{1}{4}\left \|\textbf{z}-2\textbf{x} \right \|_{2}^{2}+ \left \|\textbf{x} \right \| _{2}^{2},
\end{align}
s.t.
\begin{align}P_{S}=\{\textbf{u}_{k}|~P_{S_{1}}\cup P_{S_{2}}\cup
P_{S_{3}}\},\end{align}
where \begin{align}&P_{S_{1}}=\{\textbf{u}_{k}|~\textbf{Q}^{S_{1}}\textbf{Y}\textbf{u}_{k}\ge0 ,\quad \forall i\in S_{1}\},\\
&P_{S_{2}}=\{\textbf{u}_{k}|~\textbf{Q}^{S_{2}}\textbf{Y}\textbf{u}_{k}=0,\quad \forall i\in S_{2}\},\nonumber\\
&P_{S_{3}}=\{\textbf{u}_{k}|~\textbf{Q}^{S_{3}}\textbf{Y}\textbf{u}_{k}\le 0, \quad \forall i\in S_{3}\}.\nonumber
\end{align}

Proof: First, we analyze the one-sided dual constraint in Eq. (35) as follows 
\begin{align}
\max_{\textbf{z}:\left \| \textbf{u}_{k} \right \|_{2}\le 1}\textbf{z}^{T}\tau  \left ( \textbf{Y}\textbf{u}_{k}\right )_{\lambda }\le 2\beta.\end{align}

To divide hyperplanes, we divide $\mathbb{R}^{m^{2}}$ into three subsets to obtain Eq. (4) and Eq. (5). Let $i\in H_{1}\cup H_{2}\cup H_{3}$, $|H_{1}|+|H_{2}|+|H_{3}|=nh=I$, $\mathcal{H}_{X}$ be the set of all hyperplane arrangement patterns for the matrix $\textbf{Y}$, defined as the following set\cite{30},\cite{31}
\begin{align}\mathcal{H}_{X}=\{sign(\textbf{Y}\textbf{u}_{k}+\lambda)\cup sign(\textbf{Y}\textbf{u}_{k}-\lambda)| \textbf{u}_{k}\in{\mathbb{R}^{m^{2}}}\}.\end{align}

Next, we take out the positions of the elements corresponding to different symbols and assign them according to

\begin{align}
&S_{1}=\{i|~i\in H_{1}\}\cup \{i|~i\in H_{2}\},\\
&S_{2}=\{i|~i\in H_{2}\},\nonumber\\
&S_{3}=\{i|~i\in H_{2}\}\cup \{i|~i\in H_{3}\},\nonumber\\
&S=S_{1}\cup S_{2}\cup S_{3}.\nonumber
\end{align}

To assign a corresponding value to the position of each $i$ in the above three sets such that the same transformation as the soft threshold function is achieved, the diagonal matrix $\textbf{Q}^{S}$ is constructed, and its diagonal elements for $\textbf{Q}_{ii}$ as Eq. (8).

Using the diagonal matrix $\textbf{Q}^{S}$,  the constraints in Eq. (35) are  equivalent to the following form
\begin{align}
\max_{
\begin{matrix}
\left \| \textbf{u}_{k} \right \| _{2}\le 1
,P_{S}
\end{matrix}}|\textbf{z}^{T}\textbf{Q}^{S}\left ( \textbf{Y}\textbf{u}_{k} \right )|\le 2\beta,\end{align}
where $\textbf{Q}^{S}=\textbf{Q}^{S_{1}}+\textbf{Q}^{S_{2}}+\textbf{Q}^{S_{3}}$.

Hence, Eq. (36) can be finitely parameterized as
\begin{align}d^{\ast}_{2}=\max_{\textbf{z}}-\frac{1}{4}\left \|\textbf{z}-2\textbf{x} \right \|_{2}^{2}+ \left \|\textbf{x} \right \| _{2}^{2},\nonumber\end{align}
s.t.
\begin{align}
\max_{
\begin{matrix}
\left \| \textbf{u}_{k} \right \| _{2}\le 1
,P_{S}
\end{matrix}}|\textbf{z}^{T}\textbf{Q}^{S}\textbf{Y}\textbf{u}_{k} |\le 2\beta.\end{align}
$\hfill\qedsymbol$

Now, we introduce an exact representation of a two-layer ST-CNN.

\emph{Theorem 5}: \begin{align}d^{\ast}_{2}=\max_{\textbf{z}}-\frac{1}{4}\left \|\textbf{z}-2\textbf{x} \right \|_{2}^{2}+ \left \|\textbf{x} \right \| _{2}^{2},\nonumber\end{align}
s.t.
\begin{align}
\max_{
\begin{matrix}
\left \| \textbf{u}_{k} \right \| _{2}\le 1
,P_{S}
\end{matrix}}|\textbf{z}^{T}\textbf{Q}^{S}\textbf{Y}\textbf{u}_{k} |\le 2\beta.\nonumber\end{align}
is equivalent to
\begin{align}
d_{3}^{*}=\min_{\textbf{w}_{i}\in p_{w},\textbf{w}'_{i}\in p_{w'}}&\left \| \sum_{i=1}^{I}\textbf{Q}^{S}\textbf{Y}(\textbf{w}'_{i}-\textbf{w}_{i})-\textbf{x}\right \|_{2}^{2}\label{eq:2A}\\\nonumber
&+2\beta\sum_{i=1}^{I} (\left \| \textbf{w}_{i} \right \|_{2}+ \left \|\textbf{w}'_{i} \right \|_{2}),\nonumber
\end{align}
where,
$$p_{w}=\left \{ \textbf{w}_{i}|\textbf{Q}^{S_{1}}\textbf{Y}\textbf{w}_{i} \ge 0,\textbf{Q}^{S_{2}}\textbf{Y}\textbf{w}_{i}=0,\textbf{Q}^{S_{3}}\textbf{Y}\textbf{w}_{i}\le  0\right \},$$
$$p_{w'}=\left \{ \textbf{w}'_{i}|\textbf{Q}^{S_{1}}\textbf{Y} \textbf{w}'_{i}\ge 0, \textbf{Q}^{S_{2}}\textbf{Y}\textbf{w}'_{i}=0, \textbf{Q}^{S_{3}}\textbf{Y}\textbf{w}'_{i}\le  0\right \}.$$

The proof of Theorem 5 is provided in Appendix B. According to this theorem, we can prove that the strong duality holds, i.e. the primal ST-CNN and the dual ST-CNN achieve global optimality. They are theoretically equivalent and will obtain Theorem 6.

\emph{Theorem 6}: Suppose the optimal value of the primal ST-CNN is $p^{*}$ and the optimal value of the dual ST-CNN is $d^{*}_{3}$, the strong duality holds if $p^{*}=d^{*}_{3}$.

Proof: The optimal solution to the dual ST-CNN is the same as the optimal solution to the primal ST-CNN model constructed $ \left \{ \textbf{u}_{k}^{*},v_{k} ^{*}\right \} _{k=1}^{K}$ as follows
\begin{align}&(\textbf{u}_{k}^{*},v_{k}^{*})=(\frac{\textbf{w}_{i}^{*}}{\sqrt{\left \| \textbf{w}_{i}^{*} \right \|}},\sqrt{\left \| \textbf{w}_{i}^{*}\right \|}),\quad \text{if}\quad \textbf{w}_{i}^{*}\neq 0, \nonumber\\
&(\textbf{u}_{k}^{*},v_{k} ^{*})=(\frac{{\textbf{w}'}_{i}^{*}}{\sqrt{\left \| {\textbf{w}'}_{i}^{*} \right \|}},\sqrt{\left \| {\textbf{w}'}_{i}^{*}\right \|}),\quad \text{if}\quad {\textbf{w}'}_{i}^{*}\neq 0, 
\end{align}
where $\left \{\textbf{w}_{i}^{*},{\textbf{w}'}_{i}^{*}\right \}_{i=1}^{I}$ are the optimal solution of Eq. (45).

\begin{align}
p^{\ast} = &\min_{\textbf{u}'_{k}\in \mathbb{R}^{m},v'_{k}\in \mathbb{R}}
\left \| \sum_{k=1}^{K}\tau \left ( \textbf{Y}\textbf{u}' _{k}\right )_{\lambda } v'_{k}-\textbf{x}\right\|_{2}^{2}\\
&~~+\beta \sum_{k=1}^{K}(\left \| \textbf{u}' _{k} \right \|_{2}^{2}+|v'_{k}|^{2} )
\nonumber\\
&\le \left \| \sum_{k=1}^{K} \tau \left ( \textbf{Y}\textbf{u}^{*} _{k}\right )_{\lambda } v^{*}_{k}-\textbf{x}\right\|_{2}^{2}+\beta \sum_{k=1}^{I}(\left \| \textbf{u}^{*}_{k} \right \|_{2}^{2}+|v^{*}_{k}|^{2} )~~~~~~~~~~~~~~~~~~~~~~~~
 \nonumber\\
 &=\left \|\sum_{i=1}^{I}\textbf{Q}^{S}\textbf{Y}(\textbf{w}'_{i}-\textbf{w}_{i})-\textbf{x} \right \|_{2}^{2}\nonumber\\
 &~~+\beta  \sum_{i=1,\textbf{w}_{i}^{*}\neq 0}^{I}\left(\left \| \frac{\textbf{w}_{i}^{*}}{\sqrt{\left \| \textbf{w}_{i}^{*} \right \|_{2} } }  \right \|_{2}^{2}+\left \| \sqrt{\left \|\textbf{w}_{i}^{*} \right \|_{2} } \right \|_{2}^{2}  \right)~~~~~~~~~~~~~~~~~~~~~~~~~~~~~~~
 \nonumber\\
&~~+\beta  \sum_{i=1,\textbf{w}_{i}^{‘*}\neq 0}^{I}\left(\left \| \frac{{\textbf{w}'}_{i}^{*}}{\sqrt{\left \| {\textbf{w}'}_{i}^{*} \right \|_{2} } }  \right \|_{2}^{2}+\left \| \sqrt{\left \|\textbf{w}_{i}^{’*} \right \|_{2} } \right \|_{2}^{2} \right)~~~~~~~~~~~~~~~~~~~~~~~~~~~~~~~
 \nonumber\\
&=\left \|\sum_{i=1}^{I}\textbf{Q}^{S}\textbf{Y}(\textbf{w}_{i}^{'*}-\textbf{w}_{i}^{*})-\textbf{x} \right \|_{2}^{2}\nonumber\\
  &~~+2\beta \sum_{i=1}^{I}\left ( \left \| \textbf{w}_{i}^{*} \right \|_{2}+ \left \| \textbf{w}_{i}^{'*} \right \|_{2} \right ) = d_{3}^{\ast}. \nonumber ~~~~~~~
\end{align}

Combining $p^{*} \le d^{*}_{3} $, $p^{*} \ge d_{1}^{*}$ (Theorems 2-3) and $d_{1}^{*}=d_{2}^{*}=d_{3}^{*}$ (Theorems 4-5), $p^{*} = d^{*}_{3}$ is proved.

Basing on the Lemma 3\cite{36}, we know that $k+1$ of the total $I$ filters $(\textbf{w}_{i},\textbf{w}'_{i})$ are non-zero at optimum, where $k\le I$\cite{22,ergen2020training}.

Finally, by combining Theorems 2-6, the main theory can be proved. Thus, the hyperplane arrangements can be constructed in polynomial time (See proof in Appendix D).

The global optimization of neural networks
is NP-Hard \cite{blum1988training}. Despite the theoretical
difficulty, highly accurate models are trained in practice
using stochastic gradient methods \cite{bengio2012practical}.
Unfortunately, stochastic gradient methods cannot guarantee convergence to an
optimum of the non-convex training loss \cite{ge2015escaping} and existing methods rarely certify convergence to a
stationary point of any type \cite{goodfellow2016deep}. stochastic gradient methods
are also sensitive to hyper-parameters, they converge slowly,
to different stationary points \cite{neyshabur2017geometry} or even diverge depending on the choice of step size. Parameters like the random seed complicate replications and can
produce model churn, where networks learned using the
same procedure give different predictions for the same inputs \cite{henderson2018deep}\cite{bhojanapalli2021reproducibility}. 

Therefore, some optimizers were designed to find the optimal solution during the training process. For example, early on, the SGD optimizer \cite{robbins1951stochastic}, the SGD-based
adaptive gradient optimizer (ASGD) \cite{ASGD0} and the adoption of moment estimation (ADAM) \cite{Adam} optimizer, may lead to different training results under the same non-convex optimization objective (See results in Section IV. A)\cite{chen2018convergence}. 
\section{Experimental Results}

Experiments will show three observations: 1) The performance of the primal ST-CNN depends on the chosen optimizer. 2) The performance of the primal ST-CNN relies on initialization.  3) The zero dual gap holds between the primal and dual ST-CNN. 

All experiments were implemented on a server equipped with dual Intel Xeon Silver 4210 CPUs, 128GB RAM, the Nvidia Tesla T4 GPU (16 GB memory), and PyTorch deep learning library\cite{39}. The test dataset includes simulated data and the MNIST handwritten digits commonly used in deep learning research \cite{22}\cite{40}.

Experiments use the MNIST handwritten digits dataset with a size of 28×28. We randomly select 600 out of 60,000 as the training set, and 10,000 in the test set remain the same. Then, they are added i.i.d Gaussian noise from the distribution $\mathscr{N}(0,\sigma^{2})$ as the training and test dataset of the primal and the dual ST-CNN. 
For network training, 600 noisy images and their noise-free ones are used as the input and label. 
For the network test, 10000 noisy images and their noise-free ones are used as the input and label. 
The number of training and test datasets is consistent with that used in the ReLU-based dual theory experiments\cite{22}.
\begin{figure*}
 \centering
 \includegraphics[width=6in]{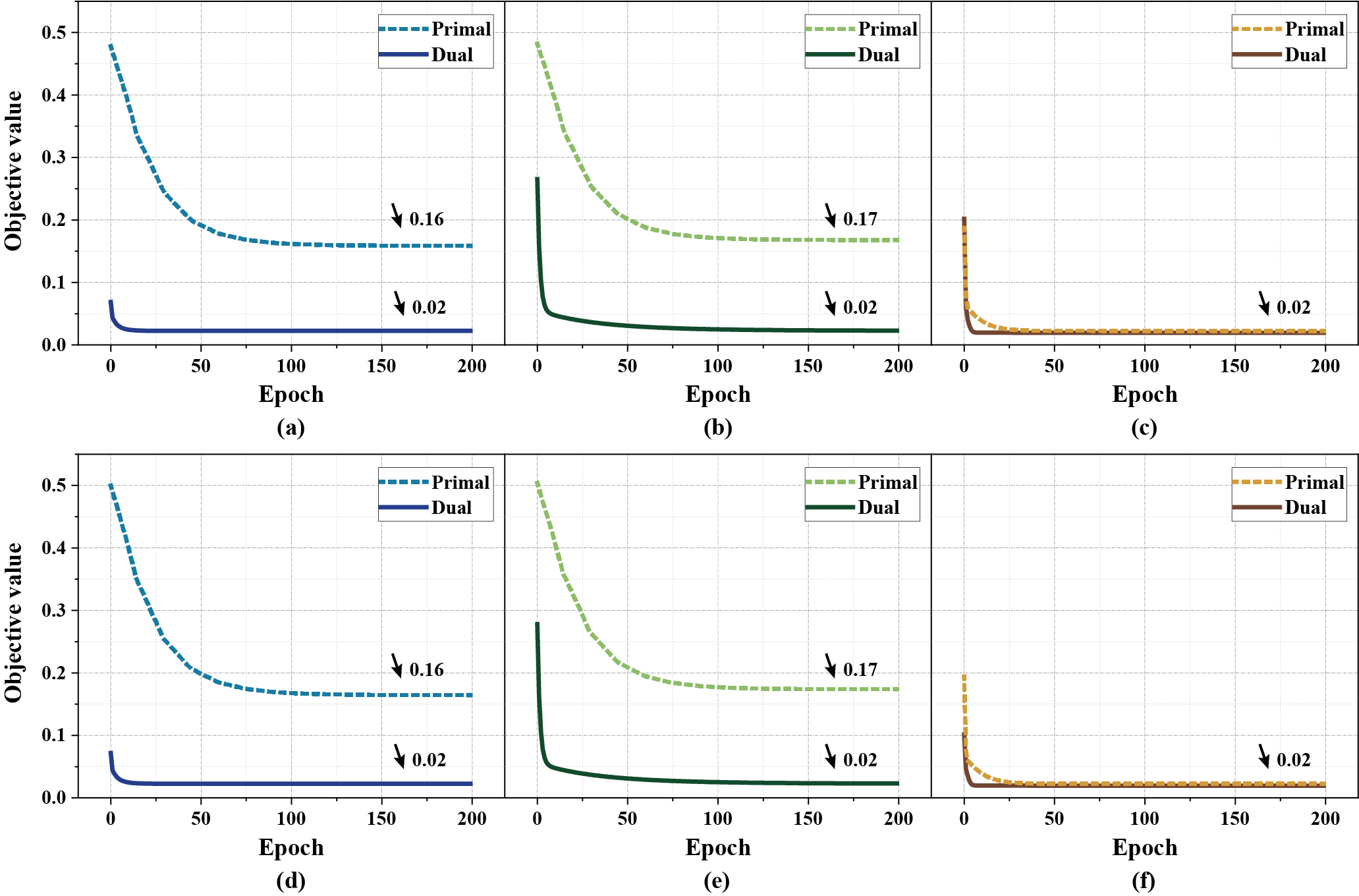}
 \caption{Training and testing with the different optimizers. From left to right: (a) Training results of the primal and primal ST-CNNs when SGD is used as an optimizer,  (b) training results when the optimizer is ASGD, (c) training results when the optimizer is ADAM, (d) testing results when the optimizer is SGD, (e) testing results when the optimizer is ASGD, (f) testing results when the optimizer is ADAM.}
\end{figure*}
\begin{figure*}
 \centering
 \includegraphics[width=6in]{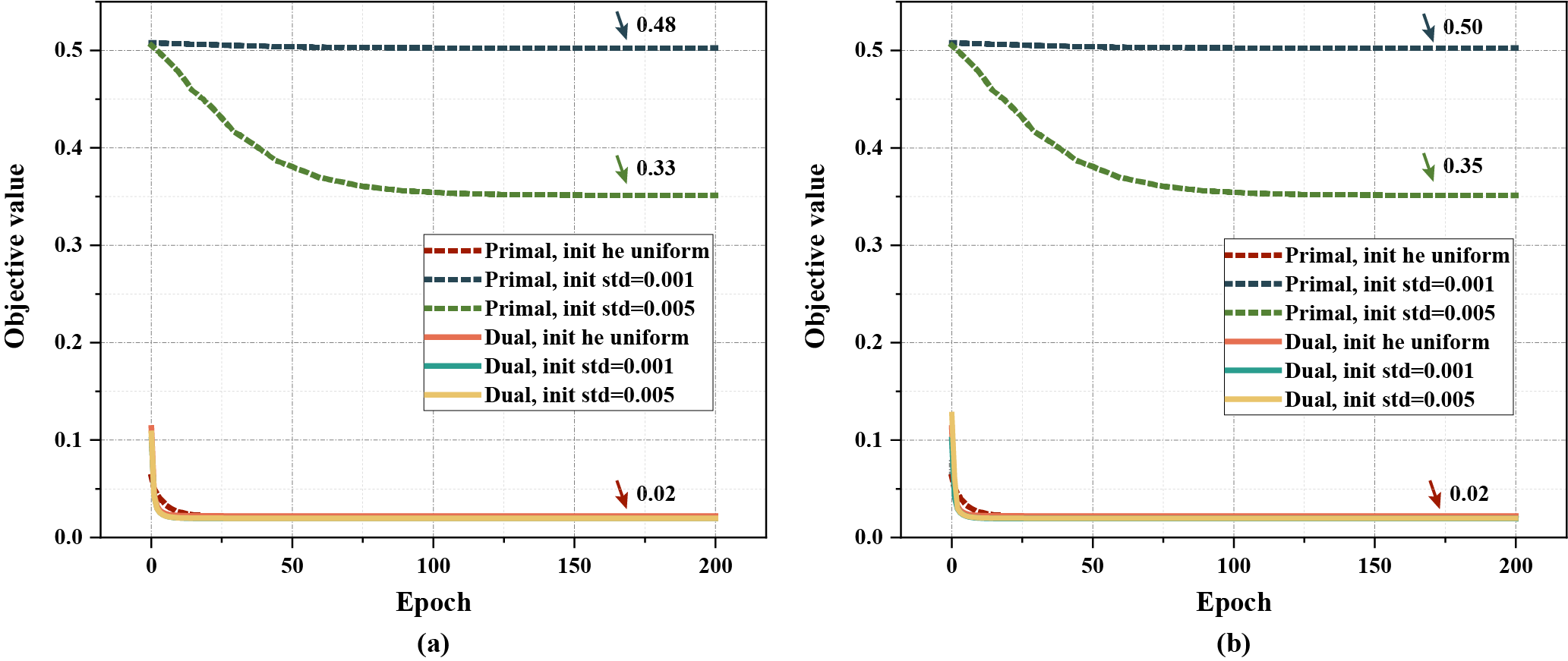}
 \caption{Example of training and testing with the different initialization. The primal ST-CNN and the dual ST-CNN are trained with Kaiming uniform initial\cite{he2015delving}, mean 0, standard deviation 0.001 and standard deviation 0.005 initialized with normal distribution, respectively. From left to right: (a) Training of the primal ST-CNN and dual ST-CNN. (b) Testing of the primal ST-CNN and dual ST-CNN. }
\end{figure*}

\begin{figure*}
   \centering
\includegraphics[width=6in]{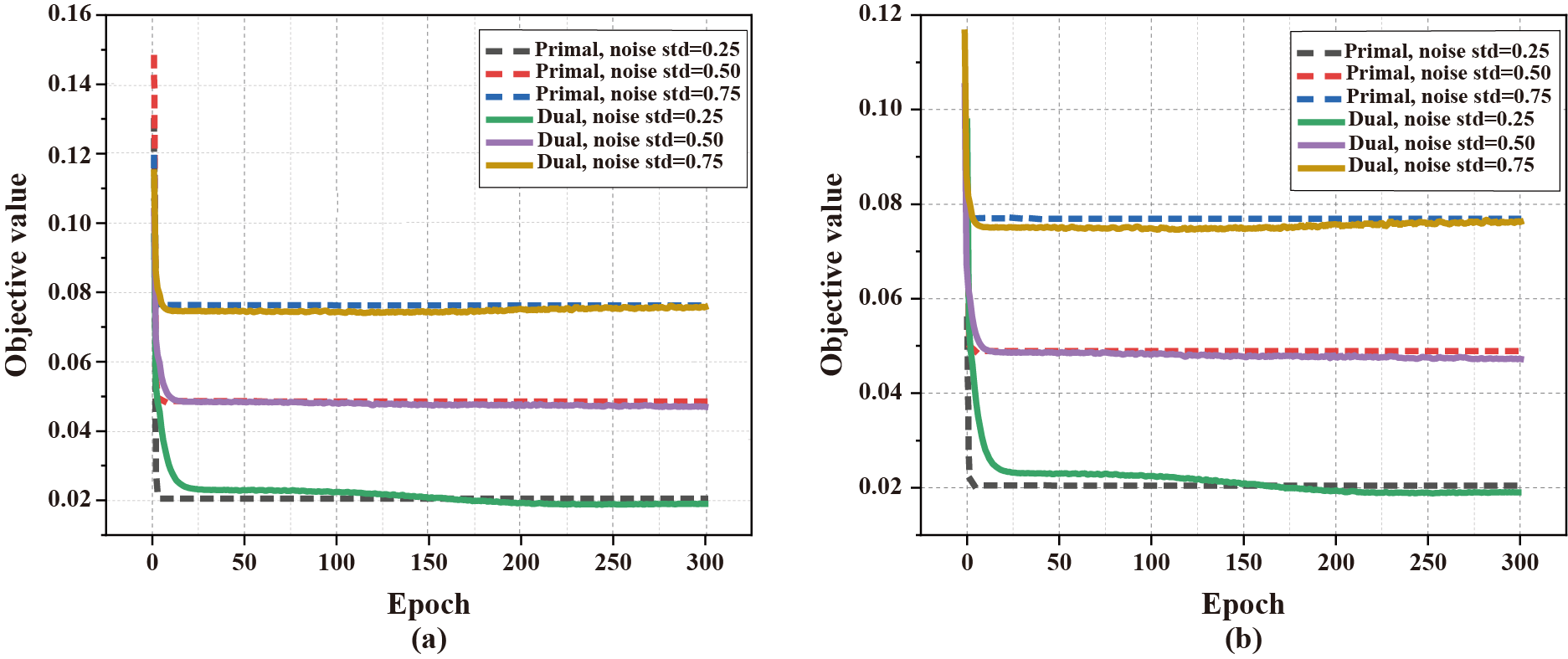}
  \caption{Verify that zero dual gap (the strong duality holds), i.e. the objective function values are very close when both the primal ST-CNN objective value and the dual 
network objective value achieves global optimality. (a) Training of the primal and dual ST-CNN in the case of Gaussian noise with mean 0 and standard deviation 0.25, 0.5, and 0.75 respectively. (b) Test of the primal and dual ST-CNNs in the case of Gaussian noise with a mean of 0 and standard deviations of 0.25, 0.5, and 0.75 respectively.}
\end{figure*}

\subsection{Primal ST-CNN Relies on Optimiser}
 
Here, we choose noise $\sigma=0.25$, the primal ST-CNN and the dual ST-CNN are trained by using SGD, ASGD\cite{ASGD0}, and ADAM\cite{Adam} as the optimizers respectively. The training and testing results are shown in Fig. 8. The optimal solution of the primal ST-CNN is dependent on the selection of optimizers. 

\subsection{Primal ST-CNN Relies on Initialization}

 We choose different ways of parameter initialization including Kaiming He uniform distribution initialized as well as a normal distribution with zero mean and standard deviation of 0.001 and 0.005, respectively. The experimental results are shown in Fig. 9. The objective value of the primal ST-CNN and the dual ST-CNN coincide when two types of networks are initialized with Kaiming He uniform distribution for training. However, when we initialize the network parameters using normal distributions with mean 0 and standard deviations of 0.001 and 0.005, the objective value of the dual ST-CNN will be better than the primal ST-CNN.

This observation implies that the primal ST-CNN is dependent on the selection of the initial values. 
\subsection{Verify Zero Dual Gap (Strong Duality)}

Zero dual gap\cite{Boyd} means that, when both the primal and the dual ST-CNN reach the global optimum, the objective values of the two are equal. Therefore, according to the above two experimental results, in order to make the primal network achieve the global optimum, we choose ADAM\cite{Adam} as the optimizer for the primal network, and the network parameters are initialized by Kaiming He uniform distribution.
 Under various noise levels $\sigma\in\{0.25, 0.5, 0.75\}$. Both approaches achieve close objective values under all noise levels (Fig. 10). 
 
 It should be noted here that the optimal values of the primal ST-CNN and the dual ST-CNN are not exactly equal as the theory proves, but very close. The error is caused by the fact that the experiment is based on a large amount of data training, which is within the negligible range. Hence, experimental results are consistent with our theory.
\section{Conclution}
In this paper, to achieve the global optimum and remove the dependence of solutions on the initial network parameters, a convex dual ST-CNN is proposed to replace its primal ST-CNN (a convolution neural network with soft-thresholding). Under the principle of convex optimal dual theory, we theoretically prove that the strong duality holds between the dual and primal ST-CNN and further verify this observation in experiments of image denoising. 

\section{Appendix}
\subsection{Proof of Theorem 2}

We combine basic inequality to rescale the parameters as
$\left \| \textbf{u}'_{k} \right \|_{2}^{2}+|v'_{k}|^{2} $ in Eq. (20)\cite{pilanci2020neural},\cite{ergen2021convex},\cite{neyshabur2014search,savarese2019infinite,ergenTolga2020convex}.

The parameters can be rescaled $\textbf{u}'_{k}=\varepsilon_{k}\textbf{u}_{k},$ and $v'_{k}=\frac{v_{k}}{\varepsilon_{k}}$, for any $\varepsilon_{k}>0.$

 \begin{align}
\sum_{k=1}^{K} \tau \left (\textbf{Y}\textbf{u}' _{k}\right ) _{\lambda} v'_{k}=\sum_{k=1}^{K} \tau  \left (\varepsilon_{k}\textbf{Y}\textbf{u} _{k}\right )_{\lambda}\frac{v_{k}}{\varepsilon_{k}}=\sum_{k=1}^{K} \tau\left (\textbf{Y}\textbf{u} _{k}\right ) _{\lambda}  v_{k}.
  \end{align}
  
This proves that the scaling has no effect on the network output.
 In addition to this, we have the following basic inequality
 \begin{align}&\min(\left \| \textbf{u}'_{k} \right \|_{2}^{2}+|v'_{k}|^{2})\\\nonumber&=\min_{\varepsilon_{k}} (\left \|\varepsilon_{k}|\textbf{u}_{k} \right \|_{2}^{2}+\left | \frac{v_{k}}{\varepsilon_{k}}\right |^{2})\\\nonumber&=2\min\left \|\textbf{u}_{k} \right \|_{2}\left |v_{k}\right|,\nonumber
  \end{align}
here $\varepsilon_{k}=(\frac{v_{k}}{\left \| \textbf{u}_{k}\right \|_{2}})^{\frac{1}{2}}.$

Since the scaling operation has no effect on the right-hand side of the inequality.
we can set $\left \| \textbf{u}_{k} \right \|_{2}=1$, $\forall k$. Therefore, $\left \|\textbf{u}_{k} \right \|_{2}\left |v_{k}\right|$ becomes $\left | v_{k} \right |$.

Now, let us consider a modified version of the problem, where the unit norm inequality constraint has no effect on the optimal solution. 
Let us assume that for a certain index $k$, $\left \| \textbf{u}_{k} \right \| _{2}\le 1$ exit $v_{k}\ne 0$ as an optimal
solution. This shows that the unit norm inequality constraint is not activated for $v_k$ and hence removing the constraint for $\textbf{u}_{k}$ will not change the optimal solution.

However, removing the constraint $\left \| \textbf{u}_{k} \right \|_{2}\longrightarrow \infty $ reduces the objective value since it yields $v_{k}=0$. 

Here, we have a contradiction that proves that all the constraints that correspond to a nonzero $v_{k}$ must be active for an optimal solution.

This also shows that replacing 
$\left \| \textbf{u}_{k} \right \| _{2}=1$ with $\left \| \textbf{u}_{k}\right \|_{2} \le 1$ is the solution to the problem.

Hence,
\begin{align}p^{\ast}=&\min_{\textbf{u}_{k}\in \mathbb{R}^{m},v_{k}\in\mathbb{R} }
\left \| \sum_{k=1}^{K} \tau  \left (\textbf{Y}\textbf{u} _{k}\right )_{\lambda} v_{k}-\textbf{x}  \right\|_{2}^{2}\nonumber\\&+2\beta \sum_{k=1}^{K}(\left \| \textbf{u} _{k} \right \|_{2}|v_{k}|).\end{align}

Relaxing the constraints without changing the optimal solution of the objective function such that 
$\left \|\textbf{u}_{k} \right \|_{2}\le1$,
so 
\begin{align}
p^{\ast}=&\min_{{\left \| \textbf{u} _{k} \right \|_{2}\le 1}}\min_{v_{k}\in\mathbb{R}}
\left \| \sum_{k=1}^{K} \tau \left ( \textbf{Y}\textbf{u} _{k}\right )_{\lambda } v_{k}-\textbf{x} \right\|_{2}^{2}\\\nonumber&+2\beta \sum_{k=1}^{K}|{v} _{k}|.\end{align}
$\hfill\qedsymbol$

\subsection{Proof of Theorem 5}

Here,
\begin{align}
\max_{
\begin{matrix}\left \| \textbf{u}_{k} \right \| _{2}\le 1,P_{S}
\end{matrix}}|\textbf{z}^{T}\textbf{Q}^{S}\left ( \textbf{Y}\textbf{u}_{k} \right ) |\le 2\beta,\end{align}
can be split into two constraints 

\begin{align}\max_{
\begin{matrix}\left \| \textbf{u}_{k} \right \| _{2}\le 1
  ,P_{S}
\end{matrix}}\textbf{z}^{T}\textbf{Q}^{S}\left ( \textbf{Y}\textbf{u}_{k} \right ) \le 2\beta,\end{align}
and 
\begin{align}
\max_{
\begin{matrix}\left \| \textbf{u}_{k} \right \| _{2}\le 1
  ,P_{S}
\end{matrix}}-\textbf{z}^{T}\textbf{Q}^{S}\left ( \textbf{Y}\textbf{u}_{k} \right )\le 2\beta.\end{align}

We first discuss the former
\begin{align}\max_{
\begin{matrix}\left \| 
\textbf{u}_{k} \right \| _{2}\le 1
,P_{S}
\end{matrix}}\textbf{z}^{T}\textbf{Q}^{S}\left ( \textbf{Y}\textbf{u}_{k} \right ) 
\Longrightarrow 
\min_{
\begin{matrix}\left \| \textbf{u}_{k} \right \| _{2}\le 1
 ,P_{S}
\end{matrix}}-\textbf{z}^{T}\textbf{Q}^{S}\left ( \textbf{Y}\textbf{u}_{k} \right ).
\end{align}

Introducing $\textbf{b},\textbf{c},\textbf{e}\in R^{I}$,
the Lagrangian form of the above function  as follows
\begin{align}L(\textbf{u}_{k})=&-\textbf{z}^{\top }\textbf{Q}^{S}\textbf{Y}\textbf{u}_{k}-\textbf{b}
\textbf{Q}^{S_{1}}\textbf{Y}\textbf{u}_{k}+\textbf{c} \textbf{Q}^{S_{2}}\textbf{Y}\textbf{u}_{k}\nonumber\\&+\textbf{e}  \textbf{Q}^{S_{3}}\textbf{Y}\textbf{u}_{k},\end{align}
where
\begin{align}&-( \textbf{z}^{\top }\textbf{Q}^{S}\textbf{Y}\textbf{u}_{k}-\textbf{b} 
\textbf{Q}^{S_{1}}\textbf{Y}\textbf{u}_{k}+\textbf{c} \textbf{Q}^{S_{2}}\textbf{Y}\textbf{u}_{k}+\textbf{e}  \textbf{Q}^{S_{3}}\textbf{Y}\textbf{u}_{k} ) \nonumber
\\\nonumber=&-( \textbf{z}^{\top }\textbf{Q}^{S}\textbf{Y}\textbf{u}_{k}+\textbf{b} 
\textbf{Q}^{S_{1}}\textbf{Y}\textbf{u}_{k}-\textbf{c} \textbf{Q}^{S_{2}}\textbf{Y}\textbf{u}_{k}-\textbf{e}  \textbf{Q}^{S_{3}}\textbf{Y}\textbf{u}_{k})
 \\\nonumber\ge& -\left \| \textbf{Y}^{\top }\textbf{Q}^{S}\textbf{z}+\textbf{Y}^{\top} \textbf{Q}^{S_{1}}\textbf{b}
- \textbf{Y}^{\top }\textbf{Q}^{S_{2}}\textbf{c}- \textbf{Y}^{\top }\textbf{Q}^{S_{3}} \textbf{e} \right \|_{2}\left \| \textbf{u}_{k} \right \| _{2}.\nonumber\end{align}

Let $\left \| \textbf{u}_{k} \right \| _{2}=1,$
\begin{align}
&inf L(\textbf{u}_{k})\nonumber\\&=-\left \| \textbf{Y}^{\top }\textbf{Q}^{S}\textbf{z}+\textbf{Y}^{\top }\textbf{Q}^{S_{1}}\textbf{b}
- \textbf{Y}^{\top }\textbf{Q}^{S_{2}}\textbf{c}- \textbf{Y}^{\top }\textbf{Q}^{S_{3}} \textbf{e} \right \|_{2}.\end{align}

Then, the dual problem is as follows
\begin{align}\min_{\begin{matrix}
 \textbf{b},\textbf{c},\textbf{e} \in \mathbb{R}^{I} 
 \\ \textbf{b},\textbf{e} \ge 0
\end{matrix}}\left \| \textbf{Y}^{\top }\textbf{Q}^{S}\textbf{z}+\textbf{Y}^{\top }\textbf{Q}^{S_{1}}\textbf{b}
- \textbf{Y}^{\top }\textbf{Q}^{S_{2}}\textbf{c}- \textbf{Y}^{\top }\textbf{Q}^{S_{3}} \textbf{e} \right \|_{2}.\end{align}

Hence, the constraints \quad
$$\max_{\textbf{z}:\left \| \textbf{u}_{k} \right \|_{2}\le 1}
\textbf{z}^{T}\tau\left ( \textbf{Y}\textbf{u}_{k}\right )_{\lambda } \le 2\beta,$$ can be equal as follows
\begin{align}
&\min_{\begin{matrix}
 \textbf{b},\textbf{c},\textbf{e} \in {\mathbb{R}^{I}}
 \\ \textbf{b},\textbf{e} \ge 0
\end{matrix}}\nonumber\\
&\left \| \textbf{Y}^{\top }\textbf{Q}^{S}\textbf{z}+\textbf{Y}^{\top }\textbf{Q}^{S_{1}}\textbf{b}
- \textbf{Y}^{\top }\textbf{Q}^{S_{2}}\textbf{c}- \textbf{Y}^{\top }\textbf{Q}^{S_{3}} \textbf{e} \right \|_{2}
\nonumber\\&\le 2\beta.
\end{align}

According to the above equation, we can deduce that 
\begin{align}&\forall i\in{[1,I]},~\exists \quad \textbf{b} _{i},\textbf{c}_{i},\textbf{e}_{i}\in R^{I},~~ s.t.~~\textbf{b}_{i},\textbf{e}_{i} \ge 0
\nonumber\\&\left \| \textbf{Y}^{\top }\textbf{Q}^{S}\textbf{z}+\textbf{Y}^{\top }\textbf{Q}^{S_{1}}\textbf{b}_{i} 
- \textbf{Y}^{\top }\textbf{Q}^{S_{2}}\textbf{c}_{i}- \textbf{Y}^{\top }\textbf{Q}^{S_{3}} \textbf{e}_{i} \right \|_{2}\nonumber\\&\le 2\beta. 
\end{align}

Considering the constraints on both sides, introducing variables\quad
$\textbf{b}'_{i},~ \textbf{c}'_{i},~ \textbf{e}'_{i}.$\quad

Then, there are $2I$ constraints as follows:
\begin{align}\max_{\begin{matrix}\textbf{z}
 \\\textbf{b} _{i},\textbf{c}_{i},\textbf{e}_{i}\in \mathbb{R}^{I}
 \\\textbf{b}_{i},\textbf{e}_{i} \ge 0
 \\\textbf{b}' _{i},\textbf{c}'_{i},\textbf{e}'_{i}\in \mathbb{R}^{I}
 \\\textbf{b}' _{i},\textbf{e}'_{i}\ge 0
\end{matrix}}-\frac{1}{4}\left \|\textbf{z}-2\textbf{x} \right \|_{2}^{2}+ \left \| \textbf{x} \right \| _{2}^{2},\end{align}
 s.t.
\begin{align} &\left\| \textbf{Y}^{\top }\textbf{Q}^{S}\textbf{z}+\textbf{Y}^{\top }\textbf{Q}^{S_{1}}\textbf{b}_{1} 
- \textbf{Y}^{\top }\textbf{Q}^{S_{2}}\textbf{c}_{1}- \textbf{Y}^{\top }\textbf{Q}^{S_{3}} \textbf{e}_{1} \right \|_{2} \nonumber\\&\le 2\beta 
 \nonumber\\&~~~~~~~~~~~~~~~~~~~~~~~~~~~~~~~~~~~~~~~~\vdots
 \nonumber\\& \left\| \textbf{Y}^{\top }\textbf{Q}^{S}\textbf{z}+\textbf{Y}^{\top }\textbf{Q}^{S_{1}}\textbf{b}_{I}
- \textbf{Y}^{\top }\textbf{Q}^{S_{2}}\textbf{c}_{I}- \textbf{Y}^{\top }\textbf{Q}^{S_{3}}\textbf{e}_{I} \right \|_{2} \nonumber\\&\le 2\beta 
 \nonumber\\&\left\|- \textbf{Y}^{\top }\textbf{Q}^{S}\textbf{z}+\textbf{Y}^{\top }\textbf{Q}^{S_{1}}\textbf{b}'_{1} 
- \textbf{Y}^{\top }\textbf{Q}^{S_{2}}\textbf{c}'_{1}- \textbf{Y}^{\top }\textbf{Q}^{S_{3}}\textbf{e}'_{1} \right \|_{2} \nonumber\\&\le 2\beta 
 \nonumber\\&~~~~~~~~~~~~~~~~~~~~~~~~~~~~~~~~~~~~~~~~\vdots
 \nonumber\\&\left\| -\textbf{Y}^{\top }\textbf{Q}^{S}\textbf{z}+\textbf{Y}^{\top }\textbf{Q}^{S_{1}}\textbf{b}'_{I}
- \textbf{Y}^{\top }\textbf{Q}^{S_{2}}\textbf{c}'_{I}- \textbf{Y}^{\top }\textbf{Q}^{S_{3}} \textbf{e}'_{I} \right \|_{2} \nonumber\\&\le 2\beta.\nonumber 
\end{align}

Noting that, as long as $\beta>0$, let $\textbf{b}_{i}=\textbf{c}_{i}=\textbf{e}_{i}=\textbf{b}'_{i}=\textbf{c}'_{i}=\textbf{e}'_{i}=\textbf{z}=\textbf{0}$, the above constraint holds, and the strong duality holds from the Slater's condition.

The dual problem can be rewritten as
\begin{align}
&\min_{\begin{matrix}\lambda_{i}, \lambda_{i}^{'} \in \mathbb{R}
 \\\lambda_{i},\lambda_{i}^{'}\ge 0 
\end{matrix}}\max_{\begin{matrix}\textbf{z}
 \\\textbf{b} _{i},\textbf{c}_{i},\textbf{e}_{i}\in \mathbb{R}^{I}
 \\\textbf{b}_{i},\textbf{e}_{i} \ge 0
 \\\textbf{b}' _{i},\textbf{c}'_{i},\textbf{e}'_{i}\in \mathbb{R}^{I}
 \\\textbf{b}'_{i},\textbf{e}'_{i}\ge 0
\end{matrix}}
\\&-\frac{1}{4}\left \|\textbf{z}-2\textbf{x} \right \|_{2}^{2}+ \left \| \textbf{x} \right \| _{2}^{2}\nonumber
 \\&+\sum_{i=1}^{I} 2\lambda_{i}\beta+\sum_{i=1}^{I} 2\lambda_{i}'\beta \nonumber
\\&-\sum_{i=1}^{I}\lambda_{i}\left\| \textbf{Y}^{\top }(\textbf{Q}^{S}\textbf{z}+\textbf{Q}^{S_{1}}\textbf{b}_{i}
-\textbf{Q}^{S_{2}}\textbf{c}_{i}-\textbf{Q}^{S_{3}} \textbf{e}_{i} )\right \|_{2} 
 \nonumber\\
 &-\sum_{i=1}^{I}\lambda_{i}'\left\| \textbf{Y}^{\top }(-\textbf{Q}^{S}\textbf{z}+\textbf{Q}^{S_{1}}\textbf{b}'_{i}
- \textbf{Q}^{S_{2}}\textbf{c}'_{i}- \textbf{Q}^{S_{3}}\textbf{e}'_{i})\right \|_{2}.\nonumber\end{align}

Introducing variable $\textbf{t}_{1},..., \textbf{t}_{I},\textbf{t}'_{1},..., \textbf{t}'_{I}\in \mathbb{R}^{m^{2}}$, the above formula can be changed to
\begin{align}
&\min_{\begin{matrix}\lambda_{i}, \lambda_{i}^{'} \in \mathbb{R}
 \\\lambda_{i}, \lambda_{i}^{'}\ge 0 
\end{matrix}}
\max_{\begin{matrix} 
\textbf{z}
 \\ \textbf{b} _{i},\textbf{c}_{i},\textbf{e}_{i}\in \mathbb{R}^{I}
 \\ \textbf{b}_{i},\textbf{e}_{i} \ge 0
 \\ \textbf{b} _{i}^{'},\textbf{c}_{i}^{'},\textbf{e}_{i}^{'}\in \mathbb{R}^{I}
 \\ \textbf{b} _{i}^{'},\textbf{e}_{i}^{'}\ge 0
\end{matrix}}
\min_{\begin{matrix}
\textbf{t}_{i}\in \mathbb{R}^{m^{2}}
  ,\left \| \textbf{t}_{i} \right \|_{2}\le 1  
\\ \textbf{t}_{i}{’}\in \mathbb{R}^{m^{2}}
 ,\left \| \textbf{t}_{i}' \right \|_{2}\le 1
 \\ \forall i\in[1,I]
 \end{matrix}}\\
&-\frac{1}{4}\left\|\textbf{z}-2\textbf{x} \right\|_{2}^{2}+ \left\| \textbf{x} \right\| _{2}^{2}\nonumber\\
&+\sum_{i=1}^{I} 2\beta(\lambda _{i}+\lambda _{i}^{'})\nonumber\\
&+\sum_{i=1}^{I}\lambda _{i}\textbf{t} _{i}^{\top }\textbf{Y}^{\top }(-\textbf{Q}^{S}\textbf{z}-\textbf{Q}^{S_{1}}\textbf{b}_{i}+\textbf{Q}^{S_{2}}\textbf{c}_{i}+\textbf{Q}^{S_{3}} \textbf{e}_{i})\nonumber\\
&+\sum_{i=1}^{I}\lambda'_{i}\textbf{t}_{i}^{'\top }\textbf{Y}^{\top }(\textbf{Q}^{S}\textbf{z}-\textbf{Q}^{S_{1}}\textbf{b}'_{i}+\textbf{Q}^{S_{2}}\textbf{c}'_{i}+\textbf{Q}^{S_{3}} \textbf{e}'_{i}).\nonumber
\end{align}

Next, we take $\textbf{z}$, $\textbf{b}_{i},\textbf{b}' _{i},\textbf{c}_{i},\textbf{c}'_{i},\textbf{e}_{i},\textbf{e}'_{i}$ as  variables to analyze the maximum value of the objective function. 

Where, we take $\textbf{z}$ as the variable to analyze the following objective functions
\begin{align}
&-\frac{1}{4}\left \| \textbf {z}-2\textbf{x} \right \| _{2}^{2}-\sum_{i=1}^{I}\lambda _{i}\textbf{t}_{i}^{\top } \textbf{Y}^{\top }\textbf{Q}^{S}\textbf{z}+\sum_{i=1}^{I}\lambda _{i}^{'}\textbf{t}_{i}^{'\top } \textbf{Y}^{\top }\textbf{Q}^{S}\textbf{z}
\nonumber\\=&-\frac{1}{4}(\left \| \textbf {z}\right \| _{2}^{2}-4\textbf{z}^{\top }\textbf{x}+4\left \| \textbf{x}\right\|_{2}^{2})-\sum_{i=1}^{I}\lambda_{i}\textbf{t}_{i}^{\top } \textbf{Y}^{\top }\textbf{Q}^{S}\textbf{z}
\nonumber\\&+\sum_{i=1}^{I}\lambda _{i}^{'}\textbf{t}_{i}^{'\top }\textbf{Y}
^{\top }\textbf{Q}^{S}\textbf{z}
\nonumber\\=&-\frac{1}{4}\left ( \left \| \textbf {z}\right \| _{2}^{2}-4\textbf{z}^{\top}\textbf{x}+4\left \| \textbf{x} \right \|_{2}^{2}+4\sum_{i=1}^{I}\lambda_{i}\textbf{t}_{i}^{\top }\textbf{Y}^{\top}\textbf{Q}^{S}\textbf{z}\right )
\nonumber\\&+\sum_{i=1}^{I}\lambda'_{i}\textbf{t}_{i}^{'\top }\textbf{Y}^{\top}\textbf{Q}^{S}\textbf{z}
\nonumber\\=&-\frac{1}{4}\left \| \sum_{i=1}^{I} \textbf{z}-(2\textbf{x}-2\lambda_{i}\textbf{Q}^{S}\textbf{Y}\textbf{t}_{i}+2\lambda'_{i}\textbf{Q}^{S}\textbf{Y}\textbf{t}_{i}^{'}) \right \|_{2}^{2}\nonumber\\
&-\left \| \textbf{x} \right \|_{2}^{2}
\nonumber\\
&+ \left \|\sum_{i=1}^{I} \textbf{x}-\lambda_{i}\textbf{Q}^{S}\textbf{Y}\textbf{t}_{i}+ \lambda'_{i}\textbf{Q}^{S}\textbf{Y}\textbf{t}'_{i}\right \|. 
\end{align}

Let $\textbf {z}=2\textbf{x}-2\lambda_{i}\textbf{Q}^{S}\textbf{Y}\textbf{t}_{i}+2\lambda'_{i}\textbf{Q}^{S}\textbf{Y}\textbf{t}'_{i}$,
the objective function can take the maximum value 
\begin{align}
&\min_{\begin{matrix}\lambda_{i}, \lambda_{i}^{'} \in \mathbb{R}
 \nonumber\\ \lambda_{i}^{'}\ge 0 
\end{matrix}}
\min_{\begin{matrix} \textbf{t}_{i}\in \mathbb{R}^{m^{2}}
  ,\left \| \textbf{t}_{i} \right \|_{2}\le 1  
 \nonumber\\ \textbf{t}'_{i}\in \mathbb{R}^{m^{2}}
 ,\left \|\textbf{t}'_{i} \right \|_{2}\le 1
  \nonumber\\ \forall i\in[1,I] \end{matrix}}\max_{\begin{matrix} 
 \textbf{b}_{i},\textbf{b}' _{i},\textbf{c}_{i},\textbf{c}'_{i},\textbf{e}_{i},\textbf{e}'_{i}\ge 0
  \nonumber\\ \textbf{b}_{i},\textbf{b}' _{i},\textbf{c}_{i},\textbf{c}'_{i},\textbf{e}_{i},\textbf{e}'_{i}\in \mathbb{R}^{I}
\end{matrix}} \nonumber\\
& \left \|\sum_{i=1}^{I}(\textbf{x}-\lambda_{i}\textbf{Q}^{S}\textbf{Y}\textbf{t}_{i}+\lambda'_{i}\textbf{Q}^{S}\textbf{Y}\textbf{t}'_{i})\right \| _{2}^{2}~~~~~~~~~~~~~~~~
  \nonumber\\
   &+\sum_{i=1}^{I}2\beta( \lambda _{i}+ \lambda'_{i})  \nonumber\\
&+\sum_{i=1}^{I}\lambda _{i}\textbf{t}_{i}^{\top }\textbf{Y}^{\top }(-\textbf{Q}^{S_{1}}\textbf{b}_{i}+\textbf{Q}^{S_{2}}\textbf{c}_{i}+\textbf{Q}^{S_{3}} \textbf{e}_{i})  \nonumber\\
  &+\sum_{i=1}^{I}\lambda'_{i}\textbf{t}_{i}^{'\top }\textbf{Y}^{\top }(-\textbf{Q}^{S_{1}}\textbf{b}'_{i}+\textbf{Q}^{S_{2}}\textbf{c}'_{i}+\textbf{Q}^{S_{3}} \textbf{e}'_{i}).
\end{align}

 Considering  $\textbf{t}_{i},..., \textbf{t}'_{i},\textbf{c} _{i},\textbf{c}' _{i},\textbf{e} _{i},\textbf{e}' _{i}$ as a variable and take the maximum value of the objective function
\begin{align}
&\min_{\begin{matrix}\lambda_{i}, \lambda_{i}^{'} \in \mathbb{R}
\nonumber \\ \lambda_{i}^{'}\ge 0 
\end{matrix}}
\min_{\begin{matrix} \textbf{t}_{i}\in \mathbb{R}^{m^{2}}
  ,\left \| \textbf{t}_{i} \right \|_{2}\le 1  
\nonumber \\ \textbf{t}'_{i}\in \mathbb{R}^{m^{2}}
 ,\left \| \textbf{t}'_{i} \right \|_{2}\le 1
\nonumber\\ \lambda_{i}\textbf{Q}^{S_{1}}\textbf{Y}\textbf{t}_{i}\ge 0
 \nonumber \\ \lambda_{i}\textbf{Q}^{S_{2}}\textbf{Y}\textbf{t}_{i}= 0
 \nonumber  \\ \lambda_{i}\textbf{Q}^{S_{3}}\textbf{Y}\textbf{t}_{i}\le 0
  \nonumber \\\lambda'_{i}\textbf{Q}^{S_{1}}\textbf{Y}\textbf{t}'_{i}\ge 0
 \nonumber \\\lambda'_{i}\textbf{Q}^{S_{2}}\textbf{Y}\textbf{t}'_{i}= 0
 \nonumber  \\\lambda'_{i}\textbf{Q}^{S_{3}}\textbf{Y}\textbf{t}'_{i}\le 0
 \nonumber \\ \forall i\in[1,I] \end{matrix}}
\nonumber \\&\left \|\sum_{i=1}^{I}(\textbf{x}-\lambda_{i}\textbf{Q}^{S}\textbf{Y}\textbf{t}_{i}
+\lambda'_{i}\textbf{Q}^{S}\textbf{Y}\textbf{t}'_{i})\right \| _{2}^{2}
\nonumber \\& +\sum_{i=1}^{I}2 \lambda _{i}\beta+\sum_{i=1}^{I}2 \lambda _{i}^{'}\beta.
\end{align}

Then, let
$\textbf{w} _i=\lambda _i\textbf{t}_i$, $\textbf{w}'_i=\lambda' _{i}\textbf{t}'_{i}$, $\textbf{t}_{i}=0$,  when $\textbf{w}_i=0$ without changing the optimal value, we obtain
\begin{align}
\min_{\begin{matrix}
\textbf{w}_{i}\in p_{w},\textbf{w}'_{i}\in p_{w'}
 \\\left \|\textbf{w} _{i} \right \|_{2}\le \lambda _{i}
 \\\left \| \textbf{w}'_{i} \right \|_{2}\le \lambda' _{i}
\\\lambda ,\lambda'\ge 0
\end {matrix}}
&\left \| \sum_{i=1}^{I} \textbf{x}-{\textbf{Q}^{S}}\textbf{Y}{\textbf{w}}_{i}
+{\textbf{Q}^{S}}{\textbf{Y}}{{\textbf{w}'} _i}\right \|_{2}^{2}\nonumber\\&+\sum_{i=1}^{I}2 \lambda _{i}\beta+\sum_{i=1}^{I} 2\lambda _{i}^{'}\beta,
 \end{align}
here,\quad $\beta >0$.
 
Let
\begin{align}\lambda_{i}=\left \| \textbf{w}_{i}  \right \|_{2},\quad
\lambda'_{i}=\left \| \textbf{w}'_{i}  \right \|_{2},\end{align}
 the objective function can get a minimum
\begin{align}
\min_{
\textbf{w}_{i}\in p_{w},\textbf{w}'_{i}\in p_{w'}} &\left \| \sum_{i=1}^{I}\textbf{x}- \textbf{Q}^{S}\textbf{Y}\textbf{w}_{i}
+\textbf{Q}^{S}\textbf{Y}\textbf{w}'_{i}\right \|_{2}^{2}\nonumber\\&+2\beta\sum_{i=1}^{I} (\left \| \textbf{w}_{i} \right \|_{2}+ \left \| \textbf{w}'_{i}\right \|_{2}).\end{align}

We further simplify to obtain
\begin{align}
\min_{\textbf{w}_{i}\in p_{w},\textbf{w}'_{i}\in p_{w'}}&\left \| \sum_{i=1}^{I}\textbf{Q}^{S}\textbf{Y}(\textbf{w}'_{i}-\textbf{w}_{i})-\textbf{x}\right \|_{2}^{2}\label{eq:2A}\\\nonumber
&+2\beta\sum_{i=1}^{I} (\left \| \textbf{w}_{i} \right \|_{2}+ \left \|\textbf{w}'_{i} \right \|_{2}).\end{align}$\hfill\qedsymbol$
\subsection{Constructing Hyperplane Arrangements in Polynomial Time}

A neuron can be expressed as $\textbf{y}_{i}^{T}\textbf{u}_{k}+\lambda$ and $\textbf{y}_{i}^{T}\textbf{u}_{k}-\lambda$
which defines a hyperplane $B_{1}=\left \{\textbf{u}_{k}|~\textbf{y}_{i}^{T}\textbf{u}_{k}+\lambda=0\right \}$, $B_{2}=\left \{\textbf{u}_{k}|~\textbf{y}_{i}^{T}\textbf{u}_{k}-\lambda=0\right \}.$
 
In addition, The regions are defined as follows
\begin{align}
&B_{1}^{+}=\left \{\textbf{u}_{k}|~\textbf{y}_{i}^{T}\textbf{u}_{k}+\lambda>0\right \},\\\nonumber
&B_{1}^{-}=\left\{\textbf{u}_{k}|~\textbf{y}_{i}^{T}\textbf{u}_{k}+\lambda<0\right\},\\\nonumber 
&B_{2}^{+}=\left \{\textbf{u}_{k}|~\textbf{y}_{i}^{T}\textbf{u}_{k}-\lambda>0\right \},\\\nonumber
&B_{2}^{-}=\left \{\textbf{u}_{k}|~\textbf{y}_{i}^{T}\textbf{u}_{k}-\lambda<0\right \}.
\end{align}

The hyperplanes of a single-layer primal ST-CNN with a two-dimensional input space, two neurons can be expressed as follows
\begin{align}
&B_{1}=\left \{ \textbf{u}_{1}|~\textbf{y}_{i}^{T}\textbf{u}_{1}+\lambda=0 \right \},\\\nonumber
&B_{2}=\left \{\textbf{u}_{1}|~\textbf{y}_{i}^{T}\textbf{u}_{1}-\lambda=0 \right \},\\\nonumber
&B_{3}=\left\{\textbf{u}_{2}|~\textbf{y}_{i}^{T}\textbf{u}_{2}+\lambda=0 \right \},\\\nonumber
&B_{4}=\left \{\textbf{u}_{2}|~\textbf{y}_{i}^{T}\textbf{u}_{2}-\lambda=0 \right \}.
\end{align}.

To investigate the number of linear regions, the following question must be answered: How many regions are generated by the arrangement of $n$ hyperplanes in $\mathbb{R}^{m^{2}}$?

According to the Lemma 4 proposed by\cite{37}, which is an extension of Zaslavsky's hyperplane arrangement theory\cite{38}. The Lemma 3 tightens this bound for a special case in which the hyperplanes may not be in general positions\cite{37}. Therefore, it is suitable for analyzing the dual ST-CNN proposed in this paper, which contains many parallel hyperplanes. Consider m hyperplanes in $\mathbb{R}^{m^{2}}$ as defined by the rows of $\textbf{Y}\textbf{u}_{k}+\lambda=0$. 

Then, the number of regions induced by the hyperplanes is as most\begin{align}\sum_{j=0}^{rank(\textbf{Y})}\left (\begin{matrix}m^{2}
 \\j
\end{matrix}  \right ).\end{align}

If the \textbf{Y} is full rank\cite{30,comput_hyperplane0, comput_hyperplane2,comput_hyperplane3}, this expression can be written as \begin{align} 3\sum_{j=0}^{r-1}\left (\begin{matrix}I-1
 \\j
\end{matrix}  \right )\le 3r\left ( \frac{e(I-1)}{r}  \right )^{r} , \end{align}
for $r\le I$, where $r:=rank(\textbf{Y})$.

It is useful to recognize that two-layer soft-thresholding networks with $K$ hidden neurons can be globally optimized via the convex program Eq. (21). The convex program has $6I^{2}$ constraints and $6Im^{2}$ variables, which can be solved in polynomial time with respect to $I$. The computational complexity is at most $O(m^{12}(\frac{I}{m^{2}})^{3m^{2}})$ using standard interior-point solvers.

\section*{Acknowledgments}
The authors thank Jian-Feng Cai, Peng Li, Zi
Wang, Yihui Huang, and Nubwimana Rachel for helpful discussions. 

\bibliographystyle{IEEEtran}
\small\bibliography{referencesv7}

\end{document}